\documentclass{article} %
\usepackage[preprint]{colm2026_conference}

\usepackage{microtype}
\usepackage{hyperref}
\usepackage{url}
\usepackage{booktabs}

\usepackage{algorithm}
\usepackage{algorithmic}
\usepackage{amsmath}
\usepackage{amssymb}
\usepackage{mathtools}
\usepackage{amsthm}
\usepackage{xspace}
\usepackage{pifont}
\usepackage{tabularx}
\usepackage{array}
\usepackage{bm}
\usepackage{colortbl}
\usepackage{wrapfig}
\usepackage{tablefootnote}
\usepackage{booktabs}
\usepackage{multicol}
\usepackage{multirow}
\usepackage{amsmath}
\usepackage{amssymb}
\usepackage{pifont}
\usepackage{graphicx}
\usepackage{tabularx}
\usepackage{array}
\usepackage{bm}
\usepackage{colortbl}
\newcolumntype{Y}{>{\raggedright\arraybackslash}X}

\usepackage{caption}
\usepackage{xcolor}

\usepackage[frozencache=true,cachedir=minted-cache]{minted} 
\definecolor{mintbg}{rgb}{0.97,0.97,0.97}
\setminted{
  bgcolor=mintbg,
  fontsize=\small,
  breaklines=true,
  frame=single
}
\usepackage[textsize=tiny]{todonotes}
\usepackage{xspace}
\newcommand{\alg}{\textsc{WayPlan}\xspace}

\usepackage{enumitem}
\usepackage{tabularx}

\usepackage{lineno}

\definecolor{darkblue}{rgb}{0, 0, 0.5}
\hypersetup{colorlinks=true, citecolor=darkblue, linkcolor=darkblue, urlcolor=darkblue}

\title{Navigating the Clutter:\\ Waypoint-Based Bi-Level Planning for Multi-Robot Systems}

\newcommand{\aspace}{\hspace{1em}}
\newcommand{\ucsb}{$^{1}$}
\newcommand{\Mit}{$^{2}$}
\newcommand{\harvard}{$^{3}$}
\newcommand{\ibm}{$^{4}$}
\newcommand{\cisco}{$^{5}$}
\author{
    Jiabao Ji\ucsb\thanks{Correspondance: <jiabaoji@ucsb.edu>} 
    \aspace 
    Yongchao Chen$^{2,3}$
    \aspace 
    \textbf{Yang Zhang}\ibm  \\
    \textbf{Ramana Rao Kompella}\cisco \aspace 
    \textbf{Chuchu Fan}\Mit \aspace
    \textbf{Gaowen Liu}\cisco \aspace
    \textbf{Shiyu Chang}\ucsb \\
    \ucsb UC Santa Barbara \aspace
    \Mit Massachusetts Institute of Technology \aspace
    \harvard Harvard University \\
    \ibm MIT-IBM Watson AI Lab \aspace
    \cisco Cisco Research
}

\let\cite\citep

\begin{document}

\ifcolmsubmission
\linenumbers
\fi

\newcommand{\boxthreedobs}{\texttt{BoxNet3D-OBS}\xspace}
\newcommand{\hlg}[2]{\setlength{\fboxsep}{0.3pt}\colorbox{green!#2}{\rule[-.05\baselineskip]{0pt}{.7\baselineskip}{#1}}}
\newcommand{\hlr}[2]{\setlength{\fboxsep}{0.3pt}\colorbox{red!#2}{\rule[-.05\baselineskip]{0pt}{.7\baselineskip}{#1}}}
\newcommand{\hlb}[2]{\setlength{\fboxsep}{0.3pt}\colorbox{aqua!#2}{\rule[-.05\baselineskip]{0pt}{.7\baselineskip}{#1}}}
\definecolor{grey}{rgb}{0.8,0.8,0.8}
\definecolor{aqua}{rgb}{0, 1, 1}
\definecolor{steel}{rgb}{0.2734, 0.5078, 0.7031}
\definecolor{slate}{rgb}{0.1836, 0.3086, 0.3086}
\definecolor{tableheadcolor}{RGB}{200,200,200}
\definecolor{transgray}{gray}{0.9}
\definecolor{lightblue}{rgb}{0.18,0.39,0.62}
\definecolor{blue2}{rgb}{0.1,0.5,0.65}
\definecolor{green2}{RGB}{0, 100, 0}
\definecolor{pink}{rgb}{0.8,0.4,0.4}
\definecolor{darkred}{RGB}{165, 42, 42}
\definecolor{lightbluebg}{rgb}{0.96,0.98,1.0}
\definecolor{rulegray}{rgb}{0.7,0.7,0.8}

\definecolor{orange2}{RGB}{221, 132, 82}
\definecolor{red2}{RGB}{196, 78, 82}
\definecolor{purple2}{RGB}{149, 108, 180}
\definecolor{darkblue2}{RGB}{76, 114, 176}
\definecolor{lightblue2}{RGB}{100, 181, 205}
\newcommand{\myrowcolour}{\rowcolor{tableheadcolor}}

\newcommand{\fullplan}{\textsc{FullPlan}\xspace}
\newcommand{\replan}{\textsc{Replan}\xspace}
\newcommand{\icreplan}{\textsc{IC-Replan}\xspace}
\newcommand{\ncreplan}{\textsc{NC-Replan}\xspace}

\newcommand{\boxnet}{\texttt{BoxNet}}
\newcommand{\boxnetsp}{\texttt{BoxNet}~}
\newcommand{\boxnettwod}{\texttt{BoxNet2D}}
\newcommand{\boxnettwodsp}{\texttt{BoxNet2D}~}
\newcommand{\boxnetthreed}{\texttt{BoxNet3D}}
\newcommand{\boxnetthreedsp}{\texttt{BoxNet3D}~}

\maketitle

\begin{abstract}
Multi-robot control in cluttered environments is a challenging problem that involves complex physical constraints, including robot–robot collisions, robot–obstacle collisions, and unreachable motions. Successful planning in such settings requires joint optimization over high-level task planning and low-level motion planning, as violations of physical constraints may arise from failures at either level. However, jointly optimizing task and motion planning is difficult due to the complex parameterization of low-level motion trajectories and the ambiguity of credit assignment across the two planning levels.
In this paper, we propose a hybrid multi-robot control framework that jointly optimizes task and motion planning. To enable effective parameterization of low-level planning, we introduce waypoints, a simple yet expressive representation for motion trajectories. To address the credit assignment challenge, we adopt a curriculum-based training strategy with a modified RLVR algorithm that propagates motion feasibility feedback from the motion planner to the task planner.
Experiments on \boxthreedobs, a challenging multi-robot benchmark with dense obstacles and up to nine robots, show that our approach consistently improves task success over motion-agnostic and VLA-based baselines.
Our code is available at \url{https://github.com/UCSB-NLP-Chang/navigate-cluster}.

\end{abstract}

\section{Introduction}\label{sec:intro}
\begin{wrapfigure}{r}{0.43\linewidth}
\small
    \vspace{-0.15in}
    \centering
    \includegraphics[width=\linewidth]{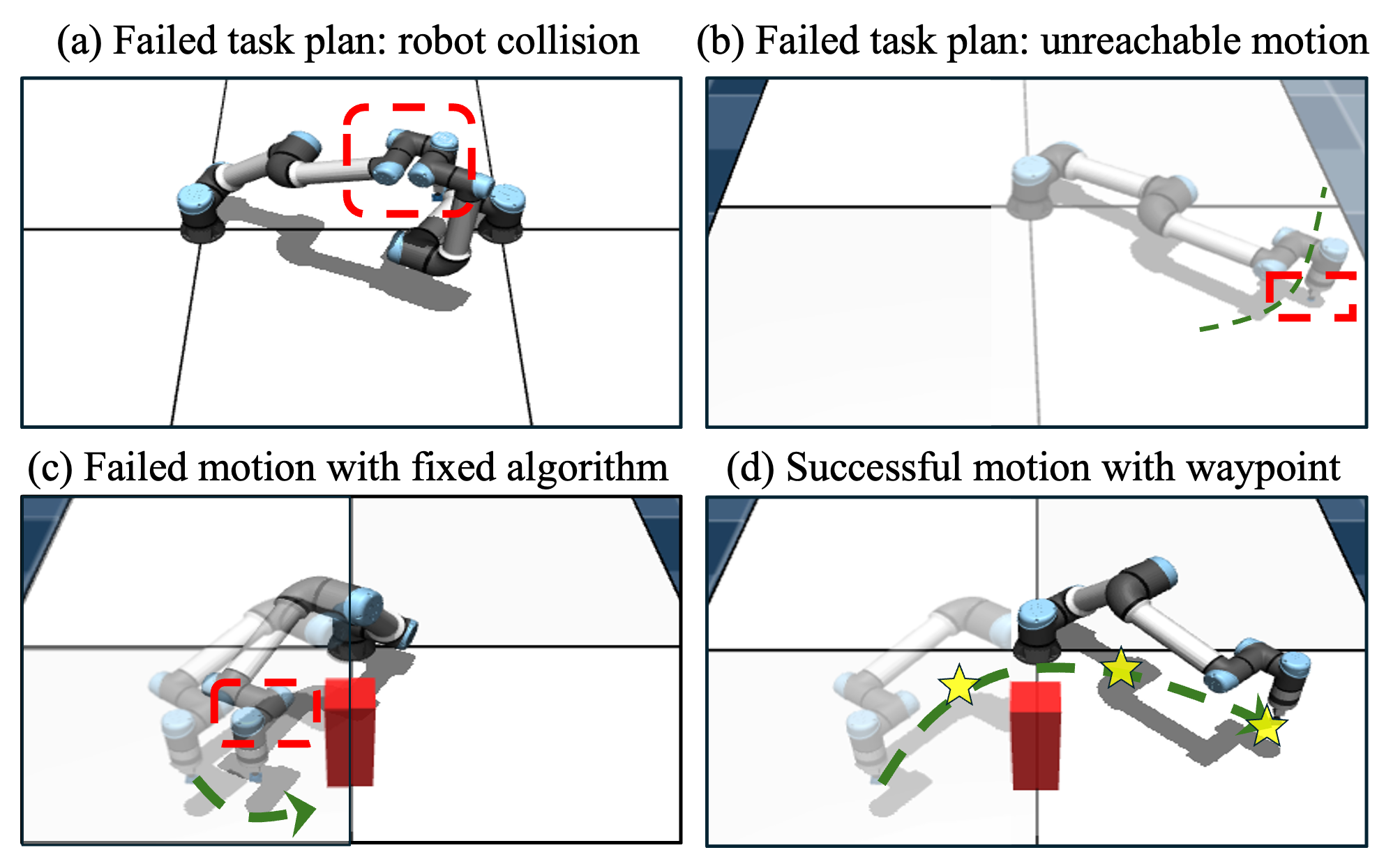}
    \vspace{-0.12in}
    \caption{Illustration of task--motion planning behaviors. (a,b): Task-level failures caused by infeasible action assignments. (c,d): Motion plans showing failure with a fixed RRT algorithm (c) and successful execution enabled by waypoint-based navigation (d).}
    \label{fig:intro-compare-motion}
    \vspace{-0.15in}
\end{wrapfigure}
Multi-robot systems are increasingly deployed in real-world domains such as warehouse automation~\cite{huang2022inner, chen2024autotamp, chen2025code0as0symbolic0planner0}, where multiple robots operate concurrently within shared workspaces.
In such settings, robots must coordinate their actions while navigating complex, obstacle-rich environments, which imposes numerous physical constraints, including kinematic feasibility, robot–obstacle collisions and inter-robot collisions.
As the number of robots and obstacles increases, interactions between robots become tightly restricted, substantially increasing the difficulty of coordinated control.

Successful task completion requires planning at two coupled levels. \textbf{Task-level planning} determines long-horizon coordination and action assignment for each robot, \textit{e.g.}, deciding which robot moves an object from its start position to a target position and in what order subtasks should be executed. \textbf{Motion-level planning} then realizes each action assignment as continuous joint configuration trajectories that avoid collisions while adhering to kinematic constraints.
Crucially, these two levels are interdependent: task assignments can lead to infeasible motion, and motion failures directly cause task execution to fail even when the high-level plan appears valid.
This coupling becomes especially severe in cluttered environments, where frequent collisions and unreachable motions arise if it is not properly addressed.

Most existing approaches address this coupling only partially. Classical methods emphasize task-level coordination, often through symbolic planning using PDDL or temporal logic for action planning, and pair it with a fixed motion planner for low-level execution~\cite{chen2024autotamp, chen2024scalable, chen2025code0as0symbolic0planner0}. More recently, large language models (LLMs) have been explored as task planners because of their strong reasoning ability, and reinforcement learning from verifiable reward (RLVR) has shown promise in improving LLMs' task plan ability~\cite{ji2025collision}. Despite this progress, the fundamental challenge of task–motion mismatch remains. 
Task-level failures occur when the high-level plan is internally infeasible, such as assigning robots to overlapping regions or unreachable positions, making success impossible regardless of the motion planner (Figure~\ref{fig:intro-compare-motion} (a) and (b)). 
Motion-level failures arise even when the task plan is valid, because a fixed or inadequate motion planner may be unable to find collision-free trajectories in cluttered environments, leading to collisions during execution (Figure~\ref{fig:intro-compare-motion} (c)). 
Together, these failure modes highlight the need for a planning approach that is \textit{jointly optimized} across both task and motion levels.

However, achieving such joint optimization in cluttered multi-robot settings is difficult. We identify two key challenges.
\textit{First}, motion-level planning is difficult to parameterize for learning. Robot motion trajectories are high-dimensional and continuous, yet must satisfy hard kinematic feasibility and collision-avoidance constraints.
Recent vision-language-action (VLA) models address motion generation by directly predicting dense joint configurations from environment observations, primarily in single-robot settings~\cite{black2410pi0}.
However, these models have been reported to suffer from unstable optimization during the RL training even in single-robot settings due to the high-dimensional outputs~\cite{huang2025corft, li2025vlarft, lyu2025flowmatchingrft}, making them difficult to scale to long-horizon, tightly constrained multi-robot environments.

\textit{Second}, credit assignment becomes inherently ambiguous when task and motion are optimized jointly. Execution failures may be caused by either a task-level error, \textit{e.g.}, allocating robots to mutually conflicting regions, or a local motion-level error, \textit{e.g.}, colliding with an obstacle.
This ambiguity makes joint RL particularly challenging, as sparse success or failure rewards must be attributed across both task- and motion-level decisions over long horizons.
These challenges lead to our key research question: \textit{How can task and motion planning be jointly optimized for multi-robot control in cluttered environments?}

In this paper, we propose \alg, a multi-robot control framework that comprises two LLM-based planners: a \textit{task planner} and a \textit{motion planner}. 
Both planners are jointly optimized for navigation in cluttered environments.
To address the first challenge in joint optimization, we introduce \textbf{waypoints}, a simplified yet expressive representation of motion trajectories.
As illustrated in Figure~\ref{fig:intro-compare-motion} (d), instead of directly predicting low-level control signals as in VLA models, the motion planner generates waypoint trajectories, \textit{i.e.} the yellow stars, that guide robots around obstacles and other robots.
These waypoints are then converted into executable robot pose trajectories using a classical motion planning algorithm such as Rapidly-exploring Random Trees (RRT)~\cite{guo2024embodied}, which delegates kinematic feasibility to a reliable planner and substantially reduces the dimensionality and data requirements of motion learning.

To address the second challenge, we introduce a \textit{curriculum training strategy} to enable stable joint optimization of task and motion planners.
We first train each planner independently using supervised fine-tuning warmup and RLVR following previous work~\cite{ji2025collision}.
We then perform joint RL training with a modified RLVR algorithm, where task-level rewards explicitly penalize plans that lead to downstream motion infeasibility, and the motion planner learns to solve feasible motion queries induced by coupled task planner.
This design allows motion failures to propagate feedback to task planner, reducing ambiguity in credit assignment between task and motion planners during joint training.

To evaluate \alg framework under realistic conditions, we introduce \boxthreedobs, a multi-robot benchmark that extends prior \textsc{BoxNet}-style environments~\cite{chen2024autotamp} with dense obstacles and challenging motion constraints.
The benchmark includes up to nine robots operating in shared grid-based layout workspaces with randomly placed obstacles, inducing frequent interactions between task-level decisions and motion feasibility.
Experiments show that \alg consistently outperforms baseline task planners.
Notably, when paired with our waypoint-based motion planner, task planners with only 4B parameters achieve a 0.62 success rate, outperforming substantially larger LLMs such as GPT-5 that lack motion awareness.
We also find that our waypoint-based motion planner outperforms VLA-style motion planner under similar data budgets.

In summary, we make the following contributions:
\begin{itemize}[
  nosep,
  topsep=0pt,
  leftmargin=1em,
  parsep=0pt,
]
\item We introduce \boxthreedobs, a multi-robot benchmark with dense obstacles and robots.
\item We propose \alg, a hybrid LLM control framework that explicitly couples task-level decisions with motion-level feasibility.
\item We develop a curriculum training strategy and a modified RLVR algorithm that enable joint optimization of task and motion planners, addressing task-motion credit ambiguity.
\item We demonstrate that waypoint motion planning outperforms VLA motion planning under similar data budgets.
\end{itemize}

\section{Method}\label{sec:method}

\vspace{-0.05in}
\subsection{\alg Overview}\label{subsec:method-overview}
We consider multi-robot control task, where multiple robots must coordinate to accomplish long-horizon tasks while respecting kinematic and collision constraints.
Our goal is to jointly reason over \emph{what} actions robots should take at each step and \emph{how} these actions can be physically executed.
To this end, we propose a hierarchical bi-level planning framework composed of two coupled planners:
\ding{182} a \emph{task planner} $\mathcal{M}_{\theta}$ that generates a multi-step action plan to assign movement target position and object carry status for involved robots, and 
\ding{183} a \emph{motion planner} $\mathcal{M}_{\phi}$ that generates obstacle-aware waypoint trajectories and connects them via a low-level control algorithm.
In the following, we first introduce the bi-level hybrid planner design in Section~\ref{subsec:hybrid-planner}, and then present our curriculum training strategy, along with the modified RLVR algorithm for joint training in Section~\ref{subsec:curriculum-train}.

\begin{figure}[!t]
    \centering
    \includegraphics[width=0.75\linewidth]{figure/hybrid-planner.png}
    \vspace{-0.06in}
    \caption{Task-level planning determines which robots act and their objectives at each step (top right), while motion-level planning generates obstacle-aware waypoint trajectories for execution (bottom right).}
    \label{fig:method}
    \vspace{-0.23in}
\end{figure}

\subsection{Bi-level Task and Motion Planning}\label{subsec:hybrid-planner}
Figure~\ref{fig:method} illustrates the planning process of \alg. Given a multi-robot task query $\bm{q}$ encoding the environment geometry, robot configurations, and the locations of objects and obstacles, \alg performs bi-level planning that integrates task-level decision making with motion-level execution.
At the task level, the planner $\mathcal{M}_{\theta}$ generates a finite-horizon plan
$\bm{s} = \{ s_1, s_2, \ldots, s_T \}$,
where each step $s_t$ specifies a task-level action for execution step $t$ (top-right panel in Figure~\ref{fig:method}).
In particular, $s_t$ identifies the subset of robots that are active at step $t$ and assigns each of them an abstract command, such as a target position and an object-carrying state.
For example, a task step may assign \emph{Robot~1 $(1.73,\,0.34,\,0.17)\ \texttt{True}$}, indicating that the robot should transport an object to the target position and release it.
Robots not selected in $s_t$ are treated as no-op and remain stationary.
This representation allows the task planner to reason over long-horizon coordination, action sequencing, and global inter-robot collision avoidance without explicitly modeling robot movement trajectory.

Each task step $s_t$ is then expanded at execution time by the motion planner $\mathcal{M}_{\phi}$.
For every robot $i$ assigned an active role in $s_t$, the motion planner produces a \textbf{waypoint-based plan}
$W_i = \mathcal{M}_{\phi}(s_t, \tau_i)$,
where $W_i = \{ w_{i,1}, \ldots, w_{i,K} \}$ is a sequence of intermediate 3D coordinates in the workspace (bottom-right panels in Figure~\ref{fig:method}).
The input $\tau_i$ captures the local perceptual state of robot $i$, including nearby obstacles, objects, and other robots within its reachable grid cells.
Rather than predicting full joint trajectories, the motion planner operates at the level of geometric waypoints, enabling efficient collision-aware spatial reasoning while remaining agnostic to robot-specific kinematics.
This abstraction significantly simplifies learning compared to VLA-based motion control.
The waypoint sequences are then executed by a low-level motion controller, such as an RRT-based planner, which converts successive waypoints into continuous, kinematically feasible robot joint trajectories.

Overall, this bi-level hybrid planner separates global coordination from local motion generation while maintaining consistency during execution.
During execution, task-level actions specify which robots act and their intended objectives, while the motion planner ensures local feasibility under geometric and collision constraints.
This decoupled design provides explicit feedback signals, including task success, collisions, and execution failures at either task or motion level, which are later leveraged for learning.

\begin{figure}[!t]
\centering
\begin{minipage}[t]{0.55\linewidth}
    \centering
    \small
    \includegraphics[width=\linewidth]{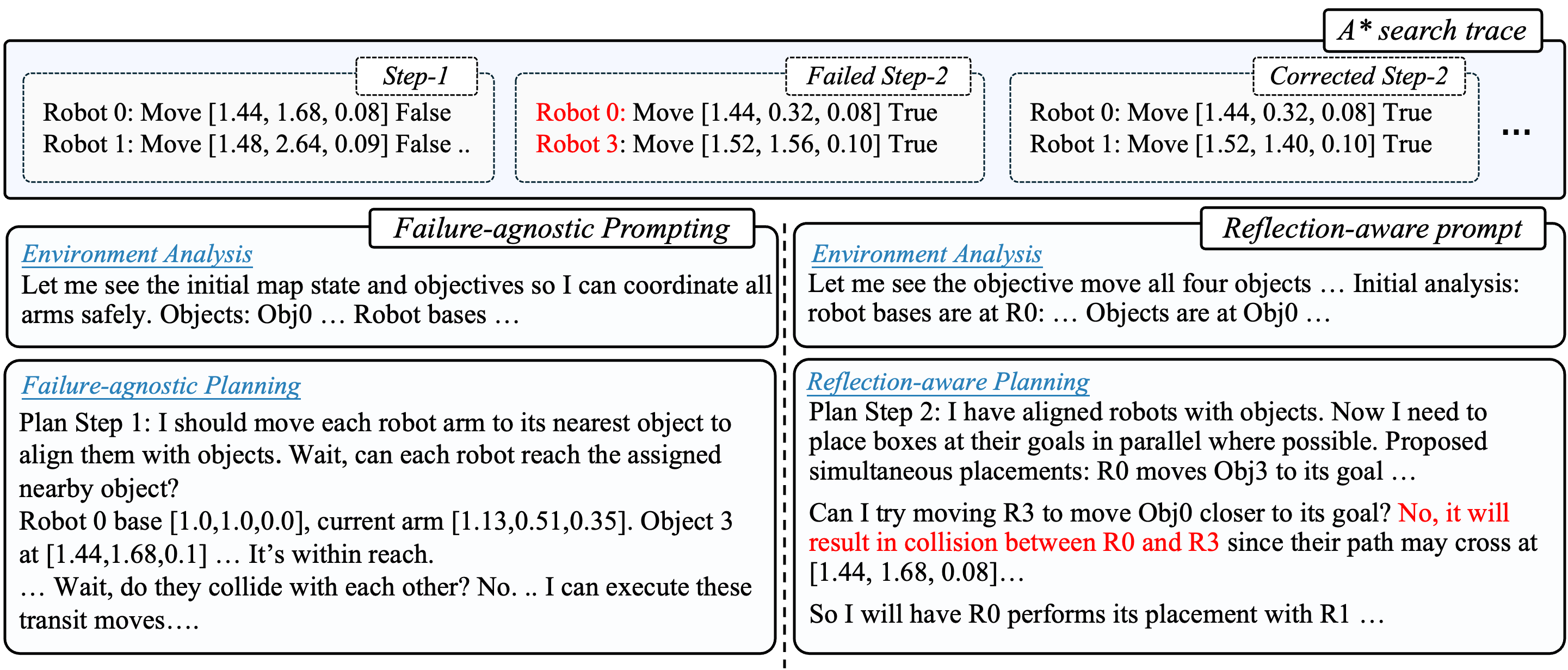}
    \vspace{-0.2in}
    \caption{Synthesized traces via failure-agnostic and reflection-aware prompting (left/right).}
    \label{fig:compare-reasoning-trace}
\end{minipage}
\hspace{0.08in}
\begin{minipage}[t]{0.35\linewidth}
    \centering
    \small
    \includegraphics[width=\linewidth]{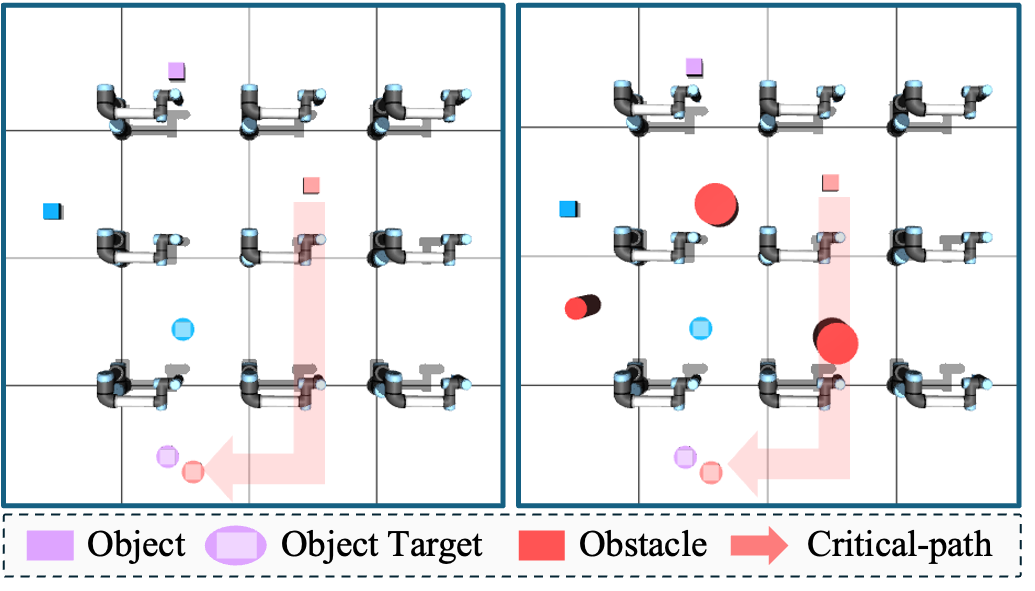}
    \vspace{-0.2in}
    \caption{Illustration of critical-path-aware obstacle placement.}
    \label{fig:boxobs-example}
\end{minipage}
\vspace{-0.18in}
\end{figure}

\subsection{Curriculum Training Strategy}\label{subsec:curriculum-train}
Directly optimizing bi-level task and motion planners is challenging, since off-the-shelf LLMs often struggle with robotic control and long-horizon reasoning~\cite{deepcoder2025, code-r1}.
We therefore adopt a three-stage curriculum training that progressively adapts LLMs for bi-level planning.
\textbf{Stage~1} performs supervised fine-tuning (SFT) warmup for both planners using planner-specific datasets, initializing basic task- and motion planning capabilities.
\textbf{Stage~2} \textit{separately} refines these two planners via separate RLVR under planner-specific reward.
\textbf{Stage~3} \textit{jointly} optimizes both planners using a modified RLVR algorithm to address task–motion coupling during joint training.

\textbf{Stage 1: SFT warm-up}\quad
In Stage~1, we initialize both the task planner and the motion planner using SFT on planner-specific data.
Because the two planners produce different outputs, we design separate SFT datasets for them.

For the task planner, each training example contains a ground-truth multi-step task plan and an explicit textual reasoning trace.
Following prior work~\cite{ji2025collision}, these traces capture long-horizon reasoning behaviors such as reachability check.
We generate these data by first solving each task instance with an A* search to obtain a ground-truth plan $\bm{s}^*$, and then prompting an LLM (GPT-5-mini) to produce a reasoning trace leading to $\bm{s}^*$.
To strengthen the reasoning, we adopt a reflection-aware construction strategy: whenever A* backtracks due to collision or reachability violations, we record the failed action, its failure reason, and the corrected alternatives.
This information is added into the prompt so the LLM can explain why the original decision is invalid and how the correction resolves the issue.
The resulting reasoning exposes common task-level planning errors and their fixes, helping task planner to reason about errors rather than simply imitating successful plans.
Figure~\ref{fig:compare-reasoning-trace} contrasts a reasoning trace between our strategy with failure-agnostic prompting of \citet{ji2025collision}, illustrating that our strategy induces error reflection reasoning (marked in \textcolor{red}{red}). Section~\ref{sec:exp-analysis} provides a more detailed analysis.

For the motion planner, we omit textual reasoning since it only involves local, short-horizon planning. Each training example comprises a collision-free waypoint sequence for a single robot navigating cluttered environments while satisfying geometric and kinematic constraints. To generate these waypoints, we sample random single-robot motion instances with randomly placed obstacles, compute ground-truth waypoint sequences via a breadth-first search (BFS) over candidate waypoints, connect them using RRT, and verify its execution feasibility in simulation.

\textbf{Stage 2: Separate RLVR with planner-specific rewards}\quad
After SFT, we further refine each planner separately using RLVR to improve their robustness while avoiding ambiguous credit assignment. We adopt standard GSPO training for both training processes.

For the task planner, it  samples candidate task plans $\bm{s} \sim \mathcal{M}_{\theta}(\bm{q})$ given a task query $\bm{q}$.
Each plan $\bm{s}$ is evaluated using the task-level reward
{
$$
r_{\text{task}}^{\text{stage-2}}(\bm{s}) =
r_{\text{success}}(\bm{s}) +
r_{\text{format}}(\bm{s}) +
r_{\text{efficiency}}(\bm{s}, \bm{s}^*).
$$
}
Here, $r_{\text{success}}(\bm{s})$ is $1$ if executing $\bm{s}$ completes the task without any collision and $0$ otherwise, $r_{\text{format}}(\bm{s}) = 0.1$ encourages the required output format, and $r_{\text{efficiency}}(\bm{s}, \bm{s}^*) = \max(0.05(|\bm{s}^*| - |\bm{s}|), -0.2)$ penalizes unnecessarily long plans relative to the A*-generated reference $\bm{s}^*$.
We intentionally omit motion penalties at Stage-2, so task-level updates depend only on task-level completion and efficiency, rather than on the quality of motion planner.

For the motion planner, we train $\mathcal{M}_{\phi}$ independently on a set of pre-generated motion tasks in a single-robot environment, without involving the task planner.
Given a motion state $(s_t,\tau_i)$, the motion planner samples waypoint sequences $W$ and receives a motion-level reward
{
$r_{\text{motion}}(W) = r_{\text{success}}(W) + r_{\text{format}}(W),$
}
where $r_{\text{success}}(W)$ is $1$ if executing $W$ moves arm to its target within a pre-set distance threshold and $0$ otherwise, and $r_{\text{format}}(W) = 0.1$ enforces waypoint format.
Both planners in Stage~2 are trained using GSPO with the rewards above, without any coupling between the task and motion levels.

\textbf{Stage 3: Joint RLVR}\quad
Stage~2 improves the task and motion planners separately, but does not ensure that task plans are matched to what the motion planner can reliably execute. Stage~3 addresses this by introducing \textit{joint RLVR}, in which the task planner is optimized with a motion-aware reward, and the motion planner is trained on motion problems generated by the current task planner.

At joint-training iteration $k$, the task planner from the previous iteration
$\mathcal{M}_{\theta}^{(k-1)}$ generates task plans
$\bm{s} \sim \mathcal{M}_{\theta}^{(k-1)}(\bm{q})$ for sampled task queries
$\bm{q}$, which are then executed with the current motion planner
$\mathcal{M}_{\phi}^{(k-1)}$.
The task-level reward augments the Stage~2 reward with a motion-feasibility
penalty,
{
$$
r_{\text{task}}^{\text{stage-3}}(\bm{s}) = r_{\text{task}}^{\text{stage-2}}(\bm{s}) +
r_{\text{motion}}(\bm{s}),
$$
where
$r_{\text{motion}}(\bm{s}) = -\max(0.2,\,0.05\times N_{\text{infeasible}}(\bm{s}))$
penalizes task plans that incur infeasible motion steps.
Here $N_{\text{infeasible}}(\bm{s})$ denotes the number of task steps whose
corresponding motion queries are certified as infeasible by invoking a
motion-planning search algorithm, \textit{i.e.}, the one for motion planner SFT data generation, under environment constraints.
This motion-aware penalty helps provide motion-feedback to help the task planner training.

The motion planner is trained on motion movement queries induced by executing task plans from the current task planner $\mathcal{M}_{\theta}^{(k-1)}$. Specifically, executing a task plan $\bm{s}$ induces a set of motion queries
$\mathcal{E}(\bm{s}) = \{(s_i, \tau_i)\}_{i=1}^N,$
where each pair corresponds to a motion execution step for one robot. We restrict training to up to $N$ motion queries that are certified as feasible by search algorithm but $\mathcal{M}_\phi^{(k-1)}$ fail, thereby isolating motion planner deficiencies rather than task planner errors. The motion planner is trained using the same motion-level reward as in Stage~2 training. 
Because these motion queries are sampled from task plans produced by current task planner $\mathcal{M}_{\theta}^{(k-1)}$, the motion planner evolves together with the task planner, tightly coupling task and motion planning and improving their alignment.
The algorithm table in Appendix~\ref{sec:app-algo-implementation} illustrates the training process.

\vspace{-0.1in}
\section{Experiment}\label{sec:experiment}
\begin{table*}[t]
\centering
\resizebox{0.7\linewidth}{!}{
\begin{tabular}{r|cc|cc|cc}
\toprule \midrule
\multirow{2}{*}{\hspace*{1.2em} \textbf{Task Planner}} &
\multicolumn{2}{c|}{\textbf{RRT Motion}} &
\multicolumn{2}{c|}{\textbf{VLA Motion}} &
\multicolumn{2}{c}{\textbf{Waypoint Motion}} \\
& {\textit{Success} $\uparrow$} & {\textit{StepDiff.} $\downarrow$}
& {\textit{Success} $\uparrow$} & {\textit{StepDiff.} $\downarrow$}
& {\textit{Success} $\uparrow$} & {\textit{StepDiff.} $\downarrow$} \\
\midrule

\rowcolor{transgray}
\multicolumn{7}{c}{LLM Task Planner without Motion-awareness (Planning in \fullplan mode)} \\
\midrule
\textsc{GPT-5-mini}       & 0.02 & 0.85 & 0.04 & 0.64  & 0.04 & 0.79 \\
\textsc{GPT-5}            & 0.04 & 0.28 & 0.11 & -0.04 & 0.14 & 0.31 \\
\textsc{Gemini3-Flash}    & 0.01 & 0.76 & 0.03 & 0.52  & 0.05 & 0.75 \\
\textsc{Gemini3-Pro}      & 0.05 & 0.24 & 0.15 & -0.11 & 0.19 & 0.12 \\
\textsc{GPT-OSS-20B}      & 0.05 & 0.31 & 0.10 & -0.01 & 0.13 & 0.25 \\
\textsc{Qwen3-4B}    & 0.02 & 1.53 & 0.03 & 1.34  & 0.02 & 1.14 \\
\textsc{Qwen3-32B}        & 0.04 & 0.48 & 0.08 & 0.26  & 0.06 & 0.43 \\
\midrule

\rowcolor{transgray}
\multicolumn{7}{c}{LLM Task Planners with Motion-awareness} \\
\midrule
{\fullplan-S1}     & 0.05 & 0.52 & 0.18 & 0.18  & 0.29 & 0.12 \\
{\fullplan-S2}      & 0.07 & 0.35 & 0.24 & -0.08 & 0.47 & -0.28 \\
{\icreplan-S1}    & 0.06 & 0.45 & 0.21 & 0.15  & 0.32 & 0.34 \\
{\icreplan-S2}     & 0.08 & 0.22 & 0.27 & \textbf{-0.12} & 0.51 & -0.12 \\
{\ncreplan-S1}    & 0.08 & 0.39 & 0.23 & 0.10  & 0.35 & 0.21 \\
{\ncreplan-S2}     & 0.09 & 0.18 & 0.29 & 0.17 & 0.53 & -0.08 \\
\midrule

\rowcolor{transgray}
\multicolumn{7}{c}{LLM Task Planners with Joint Task-Motion RLVR training} \\
\midrule
{\fullplan-S3}     & 0.08 & 0.25 & 0.26 & -0.18 & 0.53 & \textbf{-0.25} \\ 
{\icreplan-S3}    & 0.09 & 0.16 & 0.28 & -0.05 & 0.57 & -0.14 \\ 
{\ncreplan-S3}    & \textbf{0.11} & \textbf{0.14} & \textbf{0.33} & 0.09 & \textbf{0.62} & -0.11 \\ 

\midrule
\bottomrule
\end{tabular}
}
\vspace{-0.03in}
\caption{
Performance of different LLM task planner on \boxthreedobs with various motion planners.
For each task–motion planner pair, we report task-level Success and StepDiff.
Inference time is reported in Appendix~\ref{sec:app-additional-exp}.
}
\label{tab:hybrid-main}
\vspace*{-0.18in}
\end{table*}

In this section, we conduct empirical experiments on \boxthreedobs to assess the effectiveness of our method. We first present the experiment setup in Section~\ref{sec:exp-setup} and then the results in Section~\ref{sec:exp-result}, followed by ablation studies in Section~\ref{sec:exp-analysis}. Implementation details such as the data statistics and prompts are defered to Appendices~\ref{sec:app-data-implementation} and~\ref{sec:app-prompt}.

\vspace{-0.05in}
\subsection{Experiment Setup}\label{sec:exp-setup}
\textbf{Data generation}\quad
We generate \boxthreedobs environments with varying map sizes, object initial and target positions, and obstacle placements. Map sizes range from $2\times 2$ to $4\times 4$ with 1 to 6 objects, resulting in 3,400 training and 480 testing environments. Using the placement strategy described earlier, environments contain 1 to 9 robots and 1 to 8 obstacles. 
We consider only solvable environments, verified via a manually implemented A* search that ensures the existence of valid task- and motion-level plans.

\textbf{Evaluation metric}\quad
Following~\citet{ji2025collision}, we evaluate task-level planners along two dimensions: \ding{182} \textit{Success}, the proportion of generated action plans that solve a task, measured as \textit{pass@1} over four trials per environment; and \ding{183} \textit{StepDiff.}, the difference in the number of steps between successful model-generated plans and the corresponding A* solutions.
We further report the actual inference time for different planners in Appendix~\ref{sec:app-additional-exp}.

\textbf{FullPlan and RePlan mode for Task Planner}\quad
Following \citet{ji2025collision}, we consider two task-planning modes. In \fullplan, the planner generates a complete multi-step plan from the initial environment. In contrast, \replan regenerates the remaining plan after a failure, using the updated environment state and failure feedback, \textit{e.g.}, inter-robot collisions or motion failures from previous plan.
In this work, we further study two \replan variants. \icreplan appends the updated state and failure feedback to the LLM context, allowing the planner to reason over the full execution history. By contrast, \ncreplan discards prior context and replans from scratch using the updated environment as a new task instance. Illustration of reasoning for these planners are provided in Appendix~\ref{sec:app-example}.

\textbf{Baselines}\quad
For overall evaluation, we pair each task planner with three motion planner variants: \ding{182} a fixed motion algorithm (RRT Motion), \ding{183} a VLA-objective motion planner fine-tuned via SFT (VLA Motion), and \ding{184} our SFT waypoint planner (Waypoint Motion).
To ensure a fair comparison, both VLA motion and the Waypoint motion are fine-tuned on the same set of training instances only after Stage-1 training.
We also note that Stage-3 task planner evaluation pairs with Stage-3 waypoint motion planner instead of the Stage-1 checkpoint in main table.
To ensure broad LLM coverage, we evaluate both closed- and open-source models across multiple scales. Specifically, we include closed-source LLMs such as GPT-5-mini, GPT-5.2, Gemini-Flash, and Gemini3-Pro, as well as open-source Qwen3 models up to 32B parameters. To reduce API cost, we set medium reasoning effort for GPT-5-mini, GPT-5.2, and Gemini-Flash, and low for Gemini3-Pro.

\textbf{Training details}\quad
We initialize task- and motion-level planners from the same Qwen3-4B LLM~\cite{yang2025qwen3}, and denote models after each training stage with a suffix, \textit{e.g.}, –S1 for Stage-1 SFT.
In Stage-1 SFT, we train the motion planner with a learning rate of $1\times10^{-5}$ and the task planner with $3\times10^{-6}$ using the AdamW optimizer~\cite{loshchilov2017adamw}.
Stage-2 RLVR uses a fixed learning rate of $3\times10^{-6}$. We apply GRPO to the motion planner and GSPO to the task planner (with $5\times10^{-6}$ LR), using a batch size of 64 (group size 8) for both. We monitor validation reward every 10 steps and select the checkpoint where performance stabilizes; finally, we choose checkpoints around 200 and 160 steps for the motion and task planners, respectively.
We further adopt two-step off-policy RL to improve task-planner train efficiency.
Stage-3 joint RLVR training adopts the same hyperparameters as task planner Stage-2. All RL uses VeRL framework~\cite{sheng2024verl} on 8 NVIDIA H100 GPUs. Additional hyperparameters and computational cost are reported in Appendix~\ref{sec:app-algo-implementation}.

\vspace{-0.05in}
\subsection{Experiment Results}\label{sec:exp-result}
\textbf{Motion-aware Task Planners outperform motion-agnostic baselines}\quad
We first evaluate overall performance in Table~\ref{tab:hybrid-main}.
We highlight the following observations:
\ding{182} Across all three motion planners, task planners with explicit motion awareness consistently achieve higher \textit{Success} than motion-agnostic LLM planners in \fullplan mode, despite using fewer parameters.
For example, under waypoint motion, \textsc{GPT-5} and \textsc{Gemini3-Pro} achieve 0.14 and 0.19 success, respectively, whereas \fullplan-S2 achieves 0.47 success.
\ding{183} Stage-3 joint task–motion RL training further improves performance.
Specifically, \fullplan-S3 improves over its initial policy \fullplan-S2 from 0.47 to 0.53 under waypoint motion, demonstrating the benefit of jointly optimizing task and motion planners.
\ding{184} Although training is performed only with the waypoint motion planner, motion-aware task planners maintain best performance on other motion planners.
Under VLA motion, \fullplan-S3 achieves 0.26 success, outperforming baselines, whose best performance is 0.15 for \textsc{Gemini3-Pro}.
These results demonstrate that explicitly incorporating motion-feasibility during training leads to substantially improved overall performance.
To further understand the failures, we provide a detailed error breakdown in Appendix~\ref{sec:app-additional-exp}, in addition to a task-planner generalization ability evaluation.

\begin{figure}[!t]
\centering
\begin{minipage}[t]{0.5\linewidth}
    \centering
    \includegraphics[height=1.4in,width=\linewidth,keepaspectratio]{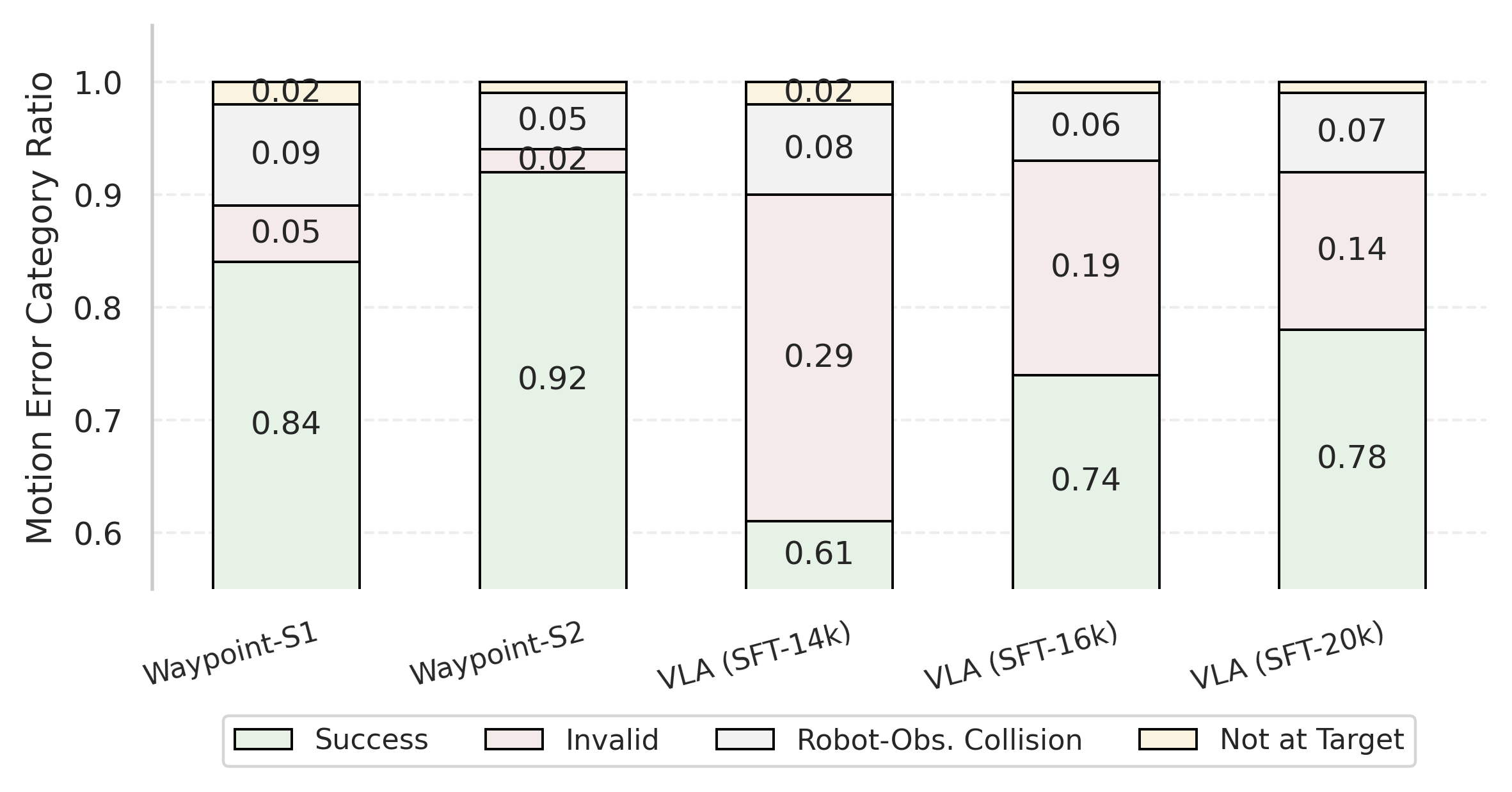}
    \vspace{-0.06in}
    \caption{Error breakdown for waypoint and VLA motion planners.}
    \label{fig:motion-errorbreak}
\end{minipage}
\hspace{0.06in}
\begin{minipage}[t]{0.45\linewidth}
    \centering
    \includegraphics[height=1.3in]{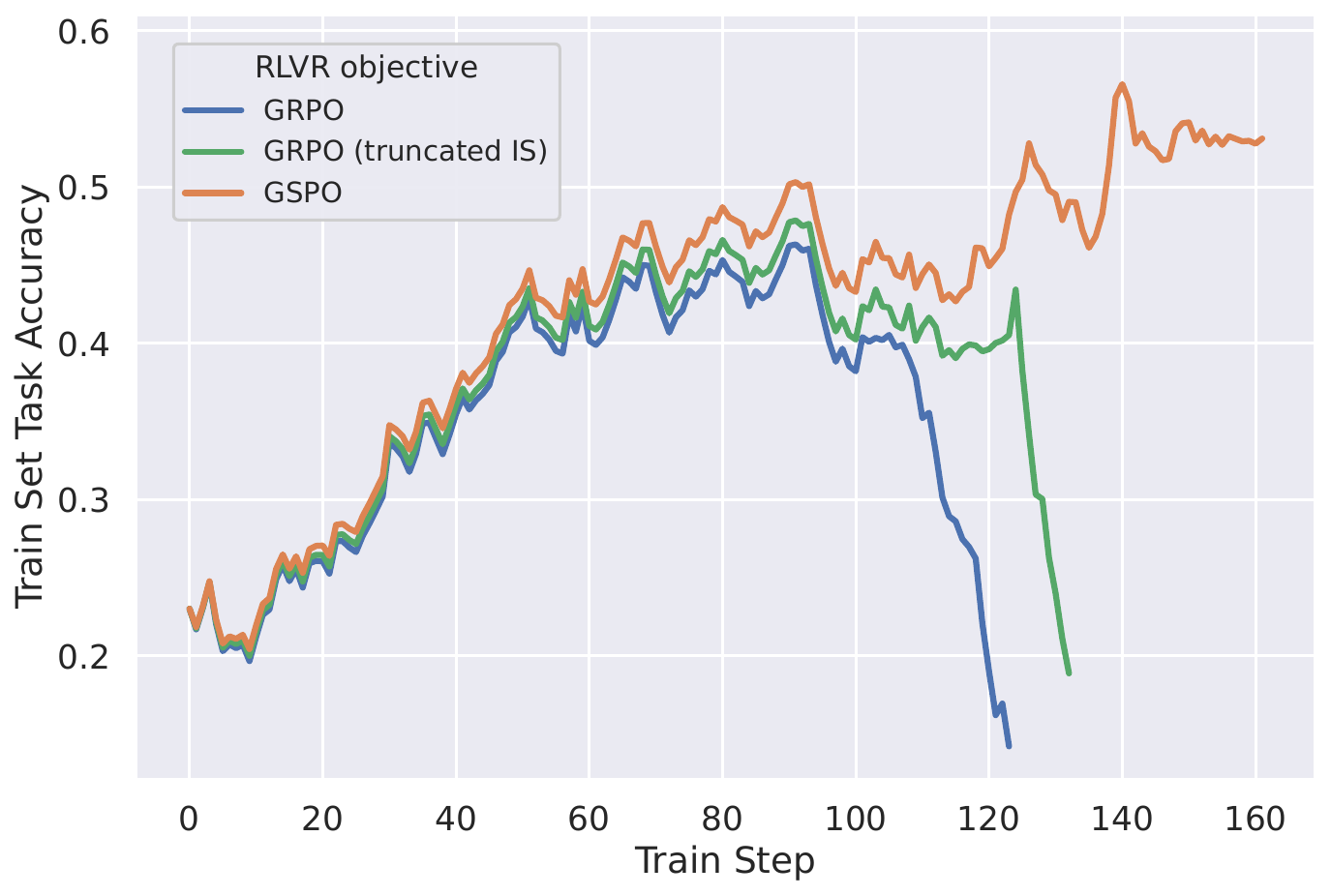}
    \vspace{-0.06in}
    \caption{Train accuracy of \fullplan-S2 across RLVR objectives. GSPO is stable.}
    \label{fig:rlvr-stability}
\end{minipage}
\vspace{-0.23in}
\end{figure}

\textbf{Replanning further improves task-level success}\quad
We next examine the effect of replanning on task-level success.
As shown in Table~\ref{tab:hybrid-main}, replanning-based planners (\icreplan and \ncreplan) consistently outperform their \fullplan counterparts when paired with strong motion planners.
Under waypoint motion, \ncreplan-S1 achieves 0.35 success, improving over \fullplan-S1 by 0.06.
This improvement becomes more pronounced after joint RL training: \ncreplan-S3 achieves 0.62 success, exceeding \fullplan-S3 by 0.09.
We also highlight that new-context replanning outperforms in-context replanning.
For example, under waypoint motion, \ncreplan-S3 achieves 0.62 success, surpassing \icreplan by 0.05.
We hypothesize this gap stems from efficient context usage, as \ncreplan avoids accumulating long execution histories, which is better suited for a 4B-parameter LLM that may falls short in long-context handling.

\textbf{Motion Planner matters in overall performance}\quad
Table~\ref{tab:hybrid-main} highlights that the choice of motion planner has a large impact on task-level success.
Across all task planners, waypoint motion consistently achieves the highest success, followed by VLA motion and then RRT motion.
Even for motion-aware task planners, performance degrades noticeably when paired with weaker motion planners, highlighting the tight coupling between task planning and motion execution.
For example, \ncreplan-S3 achieves 0.62 success with waypoint motion, but drops to 0.33 with VLA motion and further decreases under RRT motion.
These results emphasize the importance of strong motion planning.

\textbf{Waypoint Motion Planner vs. VLA Motion Planner}\quad
To better understand the performance gap between waypoint-based and VLA motion planners, we conduct an additional experiment that isolates motion execution from task-level reasoning. Specifically, we evaluate single-robot motion in cluttered $2\times2$ grids containing one robot and one obstacle. Motion movements are randomly sampled, enabling a direct assessment of motion-planner.

Figure~\ref{fig:motion-errorbreak} categorizes motion outcomes into four types: \textit{Success} (the robot successfully reaches the goal position), \textit{Invalid} (the planner produces an infeasible trajectory), \textit{Robot–Obstacle Collision} (a collision occurs during execution), and \textit{Not at Target} (the motion is feasible but fails to reach the target).
The waypoint motion planner achieves a high success rate of $0.92$ after Stage-2 training. In contrast, the VLA motion planner is dominated by \textit{Invalid}, which remain the primary source of error even as the training set scales from 14k to 20k trajectories. Although additional data improves VLA success from 0.61 to 0.78 and reduces invalid motions, its performance still falls short to waypoint planner.
We hypothesize that this gap arises from the difficulty of learning robot kinematics and collision-aware control purely from demonstrations with a 4B LLM. Unlike waypoint-based control, which explicitly enforces feasibility constraints through motion planning algorithms, VLA motion must implicitly infer these constraints, leading to persistent infeasible trajectories.

\vspace{-0.05in}
\subsection{Ablation Study}\label{sec:exp-analysis}
\begin{wraptable}{r}{0.45\linewidth}
\vspace{-0.15in}
\small
\centering

\resizebox{\linewidth}{!}{
\setlength{\tabcolsep}{4pt}
\begin{tabular}{c|cc}
\toprule \midrule
Joint Train Combinations
& {\textit{Success} $\uparrow$} & {\textit{StepDiff.} $\downarrow$} \\
\midrule
\fullplan-S2 + \texttt{Motion}-S2 & \textbf{0.53} & -0.25 \\
\fullplan-S2 + \texttt{Motion}-S1 & 0.45 & -0.12 \\
\fullplan-S1 + \texttt{Motion}-S2 & 0.37 & \textbf{-0.31} \\
\fullplan-S1 + \texttt{Motion}-S1 & 0.35 & -0.08 \\
\midrule \bottomrule
\end{tabular}
}
\captionof{table}{Task planner performance under different Stage-1/Stage-2 task and motion planner pairs for joint training.}
\label{tab:planner-combo-ablation}
\vspace{0.1in}

\resizebox{\linewidth}{!}{
\setlength{\tabcolsep}{4pt}
\begin{tabular}{c|cccc}
\toprule \midrule
SFT Data Strategy &
Reach. &
R-O Coll. &
R-R Coll. &
Err. Rec. \\
\midrule
\citet{ji2025collision}  & 7.3  & 3.9  & 4.1  & 3.7 \\
Ours  & 13.1 & 7.3  & 8.5 & 6.1 \\
\midrule \bottomrule
\end{tabular}
}
\captionof{table}{Effect of prompting strategy on reasoning behavior of \fullplan-S1.}
\label{tab:failure-step-prompt}
\vspace{-0.15in}
\end{wraptable}

\textbf{Impact of Stage-2 separate RL training}\quad
To understand whether joint task and motion training benefits from stronger individual policy initialization and which component contributes most, we evaluate the effect of Stage-2 separate RL for task and motion planners prior to joint optimization. As shown in Table~\ref{tab:planner-combo-ablation}, all variants start from the same Stage-1 \fullplan task planner and motion planner but differ in whether Stage-2 RL is applied to each component.
Initializing both planners with separate RL achieves the highest success rate, showing that Stage-2 RL provides a substantially better starting point than directly entering joint training from SFT. Moreover, Stage-2 training of the task planner has a larger impact than that of the motion planner, as pairing an RL-trained task planner with an SFT motion planner significantly outperforms the other combination in \textit{Success}, while omitting Stage-2 RL for both planners yields the weakest performance.
Overall, these results demonstrate that separate RL is crucial for both planner training and effective Stage-3 joint training.

\textbf{Impact of RL algorithm}\quad
In our initial experiments, we observe that naively applying GRPO often results in unstable training dynamics.
To better understand this behavior, we systematically examine how different RLVR objectives affect optimization stability during RL training.
Specifically, we focus on Stage~2 \fullplan RL and compare several widely used RLVR objectives: GRPO~\cite{deepseekai2025deepseekr1}, GRPO with truncated importance sampling (IS)~\cite{pang2025grpotruncate}, and GSPO~\cite{zheng2025group}.
As shown in Figure~\ref{fig:rlvr-stability}, GSPO demonstrates more stable training dynamic, achieving steady improvements in training task accuracy for up to 160 training steps without sudden drop. In contrast, GRPO exhibits severe instability, with an abrupt collapse around 110 steps, while truncated-IS GRPO delays this failure to around 120 steps. These observations are consistent with prior findings on long-horizon RL training for math reasoning~\cite{zheng2025group} and underscore the critical role of RLVR objective design in stabilizing training for complex reasoning tasks. Based on these results, we adopt GSPO for both Stage~2 and Stage~3 RL training in this work.

\textbf{Impact of reflection-aware SFT data generation strategy}\label{subsec:compare-sftdata}\quad
To examine how different SFT data generation strategies affect task-planner reasoning, we analyze the reasoning behaviors exhibited by the \fullplan-S1 planner trained on SFT data synthesized using different prompting schemes, evaluated on a subset of the test set.
As shown in Table~\ref{tab:failure-step-prompt}, failure-aware data generation substantially increases the number of reasoning related to all four categories: reachability check (Reach.), robot–obstacle collisions check (R–O Coll.), robot–robot collisions check (R–R Coll.), and explicit error recovery (Err. Rec.), indicating that the planner more frequently anticipates and reflects on potential execution failures. In contrast, data generated without failure annotations elicits limited corrective reasoning.
The prompts used by GPT-5-mini to identify and count these reasoning behaviors are in Appendix~\ref{sec:app-prompt}.

\section{Related Work}\label{sec:relatedwork}

\textbf{LLM-based multi-robot control system}\quad
Conventional multi-robot control mainly operates in task-level planning with formalizing goals and constraints with temporal logic or PDDL and performs control via constraint solvers~\cite{fox2003pddl2, emerson1990temporal}.
Recent LLM-based works either perform control primitive selection~\cite{guan2023leveraging, skreta2023errors, DBLP:conf/iclr/LoulaLDLPG0EFEC25}, control code generation~\cite{chen2025code0as0symbolic0planner0, meng2025audere0}, diffusion control~\cite{parimi2025diffusion, kim2025hierarchical} or combine with classical symbolic planning tools~\cite{chen2024autotamp, lin2023text2motion0, zhang2025llm, wang2024llmˆ}. Despite their progress, the overlook of low-level motion feasibility during task planning often yield failed task execution, especially when the environments becomes more cluttered with increasing obstacles and robots. 
Compared to these work, we propose a motion-aware LLM-based multi-robot control system that couples high-level task planning with motion execution and achieves high overall performance in cluttered environments.

\textbf{RLVR for LLM reasoning}\quad
Reinforcement learning with verifiable rewards (RLVR) has recently emerged as a powerful tool for improving the reasoning capabilities of LLMs~\cite{deepseekai2025deepseekr1, zheng2025group}.
By leveraging verifiable signals, such as answer exact-match in math~\cite{shao2024deepseekmath, Hou2025thinkprune0} and question answering~\cite{xie2026over, ji2025deepambigqa, jin2025search0r10}, or unit tests in code generation~\cite{liu2025harnessllm, he2025hardtests, code-r1}, RLVR significantly raises the LLMs' performance, enabling stronger generalization to real-world tasks.
Despite these successes in these domains, applying RLVR to embodied robot control remains challenging. For instance, prior work reports substantial difficulty when applying RLVR to VLA model training, due to the high dimensionality of robot joint spaces and the inherent complexity of learning robot kinematics, even in single-robot control settings~\cite{huang2025corft, li2025vlarft, lyu2025flowmatchingrft}.
In contrast, our work introduces a waypoint-based motion control scheme that substantially simplifies motion learning. Moreover, our modified RLVR algorithm mitigates ambiguous credit assignment between joint task planner and motion planner training.

\section{Conclusion}\label{sec:conclusion}
In this paper, we introduce \alg, a hybrid multi-robot control framework that jointly optimizes task coordination and motion feasibility in cluttered environments with complex physical constraints. By leveraging waypoint-based motion parameterization and curriculum training, our approach enables effective joint task–motion learning while reducing RL complexity and credit assignment challenges. Experiments on \boxthreedobs show that incorporating motion feasibility into task planning significantly outperforms motion-agnostic baselines, even surpassing larger general-purpose LLMs with a 4B model. These results highlight the importance of integrated task–motion planning for reliable multi-robot control.

\section*{Acknowledgement}\label{sec:acknowledgement}
Jiabao and Shiyu acknowledge the support from National Science Foundation(NSF) Grant IIS-2338252, NSF Grant IIS-2302730, the Open Philanthropy Research Award and Cisco Research Program.

\section*{LLM Usage Disclosure}\label{sec:llm-usage}
Throughout the development of this research paper, we leveraged multiple LLMs and coding-agent CLI tools, including Cursor and Codex, to support implementation and debugging via human–agent collaboration. LLMs were also used to assist in generating visualization scripts for producing illustrative figures in the experimental analysis. In addition, we utilized advanced LLMs such as GPT-5 and Gemini for grammar checking for draft writing including this paragraph. But the final version of this paper are human-only. The core research ideas and overall direction of this work were also developed solely by the authors, without direct involvement from LLMs.

\section*{Ethic Statement}\label{sec:impact}
This work advances multi-robot control by explicitly coupling task-level planning and motion-level feasibility in cluttered environments, demonstrating that jointly optimized task–motion planning can substantially improve performance over baselines. 
By introducing waypoint-based motion control scheme and a curriculum training strategy, the proposed framework reduces motion learning complexity, avoids ambiguous credit assignment, and enables smaller models to outperform much larger general-purpose LLMs, which has positive implications for practical deployment in  safety-critical domains such as warehouse robot control. 
At the same time, this work has several limitations: it relies on classical motion planners for final execution, which may struggle in highly dynamic or non-geometric constraint settings; the waypoint abstraction, while efficient, may not capture fine-grained manipulation or contact-rich behaviors; and joint training still incurs significant computational cost and assumes access to reliable simulation for RLVR. Another key limitation is that the proposed methods are evaluated only in simulation environment, 
as real-world deployment on large-scale multi-robot experimentation are difficult and costly. As a result, any real-world application would require additional validation, safety mechanisms, and human oversight.
Future work should explore extending the framework to more realistic dynamic environments and real-robot deployment.

\bibliography{colm2026_conference}
\bibliographystyle{colm2026_conference}

\appendix

\section{Appendix}\label{sec:appendix}

\subsection{\boxthreedobs Implementation Detail}\label{sec:app-data-implementation}

\paragraph{Data statistics}
For \boxthreedobs task-data implementation, we follow the same strategy as described in the main paper for map generation, robot placement and obstacle placement. Specifically, we use MuJoCo to implement the feasibility checks such as robot arm collision and object collisions. The map size ranges from $2\times2$ to $4\times4$, with robots ranges from $1$ to $9$ and the object number ranges from 1 to 6.  
For the obstacle generation, we avoid two obstacles appearing in one grid and make sure that in any $2\times2$ sub-grid, there's at most two obstacles appearing so that unreachable grid, where none robots can reach does not appear in the data to simulate warehouse environment.
For each map configuration, we randomly generate at most 50 different environments for training data and at most 5 for the test data. 
Detailed data statistics are summarized in Table~\ref{tab:boxobs-statistics}. 
Among these task instances in the train set, we generate SFT data for \fullplan, \icreplan, and \ncreplan reasoning trace with GPT-5-mini on 50\% of them and leave the other 50\% only appearing in RL stage following previous works~\cite{ji2025collision}.

For \boxthreedobs motion-data implementation, we randomly generate motion movements for a $2\times2$ single-robot grid environments, which is the basic workspace for one robot in the task-level movement. Specifically, we 

\begin{table}[h]
    \centering
    \resizebox{0.45\linewidth}{!}{
    \begin{tabular}{r|c|c}
    \toprule \midrule
    Dataset & Sample No. & Avg. Step. \\
    \midrule
    \boxthreedobs-train         & 3400 & 11.4 \\
    \boxthreedobs-test          & 480   & 10.9 \\
    \boxthreedobs-motion-train  & 14032 & - \\
    \boxthreedobs-motion-test   & 150   & - \\
    \midrule
    \bottomrule
    \end{tabular}
    }
    \caption{Dataset statistics for \boxthreedobs data.}
    \label{tab:boxobs-statistics}
\end{table}

\paragraph{Task-level A* search algorithm}
The task-level search used for plan generation is an A* search over full environment states. Each state snapshot caches box and arm poses and is scored by a heuristic that combines box-to-target distances with alignment and reachability bonuses/penalties. The planner maintains a min-heap open set keyed by priority (heuristic plus accumulated step cost), pops the best state, checks the goal condition (all boxes within tolerance of targets), and expands by generating parallel robot command sets. Action generation proposes feasible multi-robot command bundles by enumerating reachable picks/places, filtering for collision and kinematic constraints, and scoring step costs so promising parallel moves are explored first; failed command signatures are tracked to avoid repeating invalid combinations. 
Following A* search convention, our search prunes heavy search stage via g-score dominance, a coarse state signature cost bound to avoid revisiting near-duplicates, and simulation caching to skip re-evaluating the same command signature in Mujoco, which costs the most CPU time; accepted transitions are recorded so the final plan can be reconstructed by backtracking parents.

\paragraph{Motion-level BFS search algorithm}
The motion-plan frontier search used for waypoint generation is a priority-guided frontier expansion over pose nodes. The algorithm seeds a priority queue with the start node and repeatedly expands the lowest-priority path (priority is a weighted sum of joint-space distance to the target, workspace distance, and clearance penalty). 
On each expansion it first checks for a direct feasible edge to the target, otherwise it grows the path by exploring neighbour cells in Chebyshev-radius shells between the configured min/max radii, optionally biasing to wider shells when clearance is high. It enforces budgets on depth, total expansions, and stalled progress, and returns the first path that can directly connect to the target while satisfying the minimum chain length.

\subsection{\alg Implementation Detail}\label{sec:app-algo-implementation}

\paragraph{Pseudo code for Stage-3 joint RLVR training}
Algorithm~\ref{alg:stage3-joint-rlvr} illustrates the modified joint RLVR training involved in Stage-3 as we described in Section~\ref{sec:method}. Starting from the initial policy after Stage-2 seperate RLVR training, Stage-3 training jointly updates the task planner and motion planner, allowing them to adapt to each other

\begin{algorithm}[h]
\caption{Stage-3 RLVR Algorithm for Jointly Optimizing Task planner $\mathcal{M}_\theta$ and Motion Planner $\mathcal{M}_\phi$}
\label{alg:stage3-joint-rlvr}
\begin{algorithmic}[1]
\STATE \textbf{Input:} Stage-2 planners $\mathcal{M}_\theta^{(0)}, \mathcal{M}_\phi^{(0)}$, task set $\mathcal{D}_{\text{task}}$, GSPO, iterations $K$, group size $G$, cap $N$
\FOR{$k = 1,2,\dots,K$}
  \STATE Freeze rollout policies $\bar{\mathcal{M}}_\theta \leftarrow \mathcal{M}_\theta^{(k-1)}$, $\bar{\mathcal{M}}_\phi \leftarrow \mathcal{M}_\phi^{(k-1)}$
  \STATE Initialize buffer $\mathcal{B}\leftarrow\emptyset$
  \STATE Sample task query $\bm{q}\sim\mathcal{D}_{\text{task}}$
  \STATE Sample plans $\{\bm{s}^{(j)}\}_{j=1}^G \sim \bar{\mathcal{M}}_\theta(\bm{q})$
  \FOR{$j=1,\dots,G$}
    \STATE Execute $\bm{s}^{(j)}$ with $\bar{\mathcal{M}}_\phi$, obtain motion steps $\mathcal{E}_j$
    \STATE $n_j \leftarrow \sum_{(s_i,\tau_i)\in\mathcal{E}_j} \mathbb{I}[\text{infeasible}]$
    \STATE $r_j \leftarrow r_{\text{task}}^{\text{stage-2}}(\bm{s}^{(j)}) - \max(0.2,\,0.05\,n_j)$
    \STATE Add up to $N$ feasible-but-failed $(s_i,\tau_i)$ to $\mathcal{B}$
  \ENDFOR
  \STATE $\mathcal{M}_\theta^{(k)} \leftarrow \mathrm{GSPO}(\mathcal{M}_\theta^{(k-1)}, \{\bm{s}^{(j)}, r_j\}_{j=1}^G)$
  \STATE Sample $(s_t,\tau)\sim\mathcal{B}$, sample $\{W^{(j)}\}_{j=1}^G \sim \bar{\mathcal{M}}_\phi(s_t,\tau)$
  \STATE $\mathcal{M}_\phi^{(k)} \leftarrow \mathrm{GSPO}(\mathcal{M}_\phi^{(k-1)}, \{W^{(j)}, r_{\text{motion}}(W^{(j)})\}_{j=1}^G)$
\ENDFOR
\end{algorithmic}
\end{algorithm}

\paragraph{Train hyper-parameters and GPU Hour}
In this subsection, we summarize the hyperparameters used for planner training at each stage and report the corresponding H100 GPU hours in Table~\ref{tab:train-cost}. All experiments are conducted on a single 8-H100 GPU node, along with several 96-core CPU nodes for MuJoCo simulation. For checkpoint selection, we choose the model achieving the best performance on the validation set, which consists of a 20\% subset of the full \boxthreedobs test dataset.

For motion planner training, we use a learning rate of $1\times10^{-5}$ with a batch size of 256 and optimize for five epochs on the waypoint SFT dataset using the AdamW optimizer. In Stage-2 separate RLVR training, we apply on-policy GRPO with a learning rate of $3\times10^{-6}$ and a batch size of 64 (group size 8). The same learning rate is used in Stage-3 joint training for the motion planner, where the RLVR objective is switched to GSPO, consistent with the task planner.

For VLA motion planner training, we follow prior work and fine-tune the model to predict chunked robot joint configurations using a next-token cross-entropy loss. Each forward pass predicts the next 20 joint configurations, corresponding to 20 Hz control signal generation. To ensure fair comparison, the SFT dataset uses the same set of motion task instances as the waypoint motion planner training. For each task instance, we simulate the robot in MuJoCo and collect joint configurations at a 20 Hz sampling rate to construct the VLA SFT dataset. We employ the same set of hyper-parameters for VLA SFT.

For task planner training, we adopt a learning rate of $2\times10^{-5}$ with a batch size of 64 and train for eight epochs on the task planner SFT data using AdamW. In Stage-2 separate RL training, we use GSPO with a learning rate of $5\times10^{-6}$ and a batch size of 64 (group size 8). This learning rate is maintained in Stage-3 joint training. To improve training efficiency, we follow prevoius work and employ a two-step off-policy RL strategy for the task planner in both Stage-2 and Stage-3, where task plan data may be generated by policies up to two steps older than the current task policy.

\begin{table}[h]
\small
\centering
\vspace{-0.05in}
\resizebox{0.4\linewidth}{!}{
\begin{tabular}{r|cc}
\toprule \midrule
{Model}
& \textsc{Train Steps}
& \textsc{H100 GPU Hours} \\
\midrule
\rowcolor{transgray}
\multicolumn{3}{c}{Stage-1: SFT Models} \\ \midrule
Motion planner & 275 & 4.4 \\
{\fullplan}   & 325 & 6.8 \\
{\icreplan}   & 540 & 12.4 \\
{\ncreplan}   & 404 & 11.2 \\
\midrule
\rowcolor{transgray}
\multicolumn{3}{c}{Stage-2: Separate RLVR} \\ \midrule
Motion planner & 200 & 109.6 \\
{\fullplan}   & 160 & 182.4 \\
{\icreplan}   & 160 & 307.2 \\
{\ncreplan}   & 160 & 264.8 \\
\midrule
\rowcolor{transgray}
\multicolumn{3}{c}{Stage-3: Joint Task-Motion RLVR} \\ \midrule
{\fullplan}   & 80 & 149.6 \\
{\icreplan}   & 80 & 263.2 \\
{\ncreplan}   & 80 & 219.2 \\
\midrule
\bottomrule
\end{tabular}
}
\caption{Training computation cost across different training stages.}
\label{tab:train-cost}
\vspace{-0.15in}
\end{table}

\subsection{Additional Experiment Results}\label{sec:app-additional-exp}

\paragraph{Planner Inference Time Cost}
Latency is a critical factor for practical deployment in robot control, especially in settings where planning must be performed online. To evaluate inference-time cost, we benchmarked several planners on 50 \boxthreedobs tasks, each instantiated in a $3 \times 3$ grid environment with two obstacles. For each task, we measured both the time required to generate a feasible plan and the total execution time to complete the task simulation in MuJoCo; together, these metrics capture the end-to-end inference cost under realistic conditions. Planner serving was performed using two GPUs, with one dedicated to the task planner and the other to the motion planner. All available optimizations were enabled for the A* search.

Table~\ref{tab:latency} reports the averaged latency in seconds. 
We highlight the following observations.
First, LLM-based planners are substantially more efficient than search-based planning, reducing total runtime from 580.3 seconds for A* to under 60 seconds across all LLM-based methods. This efficiency arises because LLM-based planning avoids repeated simulation during search and instead relies on a small number of forward model calls.
Second, waypoint motion planning consistently achieves lower inference latency than VLA motion planning, as it requires only a single waypoint sequence generation per robot motion, whereas VLA planning may involve multiple LLM calls to refine low-level actions.
These results demonstrate that LLM-based planning, particularly when combined with waypoint motion planning, has potential real-world application for robot control.

\begin{table}[h]
\centering
\resizebox{0.45\linewidth}{!}{
\begin{tabular}{c|c|c|c}
\toprule \midrule
Task Planner & Solution Gen. & Simulation & Total \\
\midrule
\rowcolor{transgray} 
\multicolumn{4}{c}{Search-based method} \\
A* search & 248.4 & 331.9 & 580.3 \\ \midrule

\rowcolor{transgray}
\multicolumn{4}{c}{VLA Motion Planner} \\ \midrule

\fullplan-S3 & 22.4 & 3.5 & 25.9 \\
\icreplan-S3 & 47.7 & 6.8 & 54.5 \\
\ncreplan-S3 & 51.2 & 7.3 & 58.5 \\

\midrule
\rowcolor{transgray}
\multicolumn{4}{c}{Waypoint Motion Planner} \\ \midrule

\fullplan-S3 & 20.4 & 3.4 & 23.8 \\
\icreplan-S3 & 37.1 & 6.1 & 43.2 \\
\ncreplan-S3 & 42.4 & 6.7 & 49.1 \\

\midrule
\bottomrule
\end{tabular}
}
\caption{Inference time comparison on \boxthreedobs task.
We report average solution generation time, MuJoCo simulation time, and total end-to-end execution time (in seconds).
}
\label{tab:latency}
\end{table}

\paragraph{Task error breakdown}
To better understand the failures of different task planners, we analyze the failure modes of our task-level planner when paired with either the VLA motion planner or the waypoint-based motion planner. 

We categorize failures into motion-related failures and task-level failures.
Motion-related failures capture errors arising during motion execution. These include execution failures, where the motion planner generates infeasible trajectories; robot–obstacle collisions, where the executed trajectory results in a collision between a robot and an obstacle; robot–robot collisions, where trajectories lead to collisions between multiple robots; and far-from-target failures, where the robot terminates execution significantly away from the intended target position.
Task-level failures capture errors in high-level planning and feasibility. These include unreachable motion steps, where a planned motion step is verified as infeasible by the motion search algorithm, and task incompletion failures, where not all required objects are successfully moved to their target locations after executing all planned steps.

The results are presented in Table~\ref{tab:failure-breakdown}, .
We highlight following observations:
First, when paired with the VLA motion planner, the dominant performance bottleneck is motion execution failure, which accounts for the majority of task failures. This aligns with observations in the main paper, highlighting the relatively weaker robustness of the VLA motion planner even in single-robot settings, which becomes further amplified in multi-robot scenarios.
Second, when using waypoint-based motion, the primary limitation shifts to unreachable motion steps, indicating that task-level motion feasibility rather than execution robustness becomes the dominant challenge. This suggests that, despite improved low-level motion reliability, long-horizon multi-robot coordination remains inherently difficult, particularly in collaborative tasks requiring precise action sequencing and inter-robot dependency management.

\begin{table}[h]
\small
\centering
\resizebox{0.7\linewidth}{!}{
\begin{tabular}{r|cccc|cc}
\toprule \midrule
\multirow{2}{*}{Task Planner}
& \multicolumn{4}{c|}{\textsc{Motion-level Failure}}
& \multicolumn{2}{c}{\textsc{Task-level Failure}} \\
\cmidrule(lr){2-5}\cmidrule(lr){6-7}
& Execution err. & Rob-obs Coll. & Rob-rob Coll.
& Far from target & Unreachable motion & Far from target \\
\midrule
\rowcolor{transgray}
\multicolumn{7}{c}{VLA Motion} \\ \midrule
{\fullplan-S3} & 0.45 & 0.04 & 0.05 & 0.01 & 0.13 & 0.00 \\
{\icreplan-S3} & 0.35 & 0.09 & 0.07 & 0.03 & 0.11 & 0.01 \\
{\ncreplan-S3} & 0.31 & 0.08 & 0.09 & 0.02 & 0.15 & 0.02 \\
\midrule
\rowcolor{transgray}
\multicolumn{7}{c}{Waypoint Motion} \\ \midrule
{\fullplan-S3} & 0.02 & 0.06 & 0.09 & 0.07 & 0.20 & 0.00 \\
{\icreplan-S3} & 0.02 & 0.08 & 0.08 & 0.05 & 0.18 & 0.01 \\
{\ncreplan-S3} & 0.01 & 0.05 & 0.06 & 0.04 & 0.17 & 0.02 \\
\midrule
\bottomrule
\end{tabular}
}
\caption{Task planner failure breakdown under Motion-level failures and Task-level failures.}
\label{tab:failure-breakdown}
\vspace{-0.05in}
\vspace{-0.15in}
\end{table}

\paragraph{Task planner generalization ability}
Following prior work~\cite{ji2025collision}, we evaluate the generalization ability of task planners. We consider two generalization settings. In \textsc{UnseenMap}, the map sizes are not observed during training, specifically $2\times5$ and $8\times2$ grids . In \textsc{UnseenLayout}, robot bases are no longer fully aligned with grid joints on $3\times3$ and $4\times4$ grids, with some base locations removed from the standard layout.
For each map size, we generate 50 test instances using the same procedural generation rules as the \boxthreedobs test set. The results are reported in Table~\ref{tab:transfer-exp}, where we measure both task success rate and the step difference relative to A* searched plans.
We highlight the following observations:
\ding{182} Across both \textsc{UnseenMap} and \textsc{UnseenLayout}, RLVR models (Stage-2 and Stage-3) consistently outperform SFT models in terms of success rate and StepDiff, indicating improved generalization to unseen environment variations. This aligns with the observation in prevoius works~\cite{ji2025collision}.
\ding{183} Among all methods, {\ncreplan} achieves the strongest generalization performance, especially under joint RLVR training, attaining the highest success rates in both settings. This further confirms that \ncreplan is a strong planning mode for such complex reasoning task.

\begin{table}[h]
\small
\centering
\resizebox{0.45\linewidth}{!}{
\begin{tabular}{r|cc|cc}
\toprule \midrule 
\multirow{2}{*}{\hspace*{-4em}{Model}}  
& \multicolumn{2}{c|}{\textsc{UnseenMap}} 
& \multicolumn{2}{c}{\textsc{UnseenLayout}} \\
& {\textit{Success} $\uparrow$} & {\textit{StepDiff.} $\downarrow$} 
& {\textit{Success} $\uparrow$} & {\textit{StepDiff.} $\downarrow$} \\ 
\midrule
\rowcolor{transgray}
\multicolumn{5}{c}{Stage-1: SFT Models} \\ \midrule
    {\fullplan} & 0.17 & 0.38 & 0.15 & 0.24 \\ 
    {\icreplan} & 0.27 & 0.74 & 0.24 & 0.55 \\ 
    {\ncreplan} & 0.31 & 0.62 & 0.28 & 0.61 \\
\midrule
\rowcolor{transgray}
\multicolumn{5}{c}{Stage-2: Isolated RLVR Models} \\ \midrule
    {\fullplan}  & 0.39 & 0.21 & 0.31 & 0.13 \\ 
    {\icreplan}  & 0.46 & -0.04 & 0.37 & 0.24 \\ 
    {\ncreplan}  & 0.47 & 0.05 & 0.43 & 0.08 \\
\midrule
\rowcolor{transgray}
\multicolumn{5}{c}{Stage-3: Joint RLVR Models} \\ \midrule
    {\fullplan}  & 0.43 & 0.11 & 0.35 & -0.03 \\ 
    {\icreplan}  & 0.51 & 0.07 & 0.44 & 0.17 \\ 
    {\ncreplan}  & 0.55 & -0.08 & 0.49 & 0.11 \\
\midrule
\bottomrule
\end{tabular}
}
\caption{Task planner generalization on unseen \boxthreedobs environments.}
\label{tab:transfer-exp}
\vspace{-0.05in}
\vspace{-0.15in}
\end{table}

\subsection{Planner Mode Illustration}\label{sec:app-example}
In this section, we present example task-planner reasoning for our three different planner modes.

\paragraph{\fullplan Reasoning Trace Illustration}
As described in the main paper, \fullplan planning takes the initial environment observation as input and directly generates a multi-step movement plan. Each step specifies the actions of all involved robots, including their target positions and box-carrying states.
Table~\ref{tab:example-trace-fullplan} below presents an example reasoning trace for the given environment.

\paragraph{\icreplan Reasoning Trace Illustration}
As described in the main paper, \icreplan augments the planner’s input by appending the environment observation immediately preceding the last failed action, along with the corresponding failure feedback. This allows the planner to reason over the complete planning history when generating the remaining plan.
Table~\ref{tab:example-trace-icreplan} below presents an example reasoning trace.

\paragraph{\ncreplan Reasoning Trace Illustration}
As described in the main paper, \ncreplan refreshes its context when an action plan fails at an intermediate step. It initializes a new context containing the environment observation before the failed action, together with the failure feedback, enabling the planner to reason over the full planning history when generating the remaining plan.
Table~\ref{tab:example-trace-ncreplan} below presents an example reasoning trace.

\begin{table}[h]
\small
\centering
\begin{tabularx}{0.65\textwidth}{Y}
\toprule \midrule
\textbf{Input}: \\
\textcolor{gray}{\textless observation\textgreater} \\
\textcolor{gray}{Object positions:} \\
\textcolor{gray}{\hspace*{2em}Object 0: [0.75, 1.75, 0.08] $\dots$} \\
\textcolor{gray}{\hspace*{2em}Object 1: [2.25, 1.75, 0.08] $\dots$} \\
\textcolor{gray}{Robot states:} \\
\textcolor{gray}{\hspace*{2em}Robot 0: base=[0.25, 0.75, 0.00], arm=[0.35, 0.85, 0.10] $\dots$} \\
\textcolor{gray}{\hspace*{2em}Robot 1: base=[2.75, 0.75, 0.00], arm=[2.65, 0.85, 0.10] $\dots$} \\
\textcolor{gray}{\textless /observation\textgreater}\\[0.25em]

\textbf{\fullplan Planner Reasoning}: \\
\textcolor{blue2}{\textless think\textgreater Okay, let me analyze the given environment before coming up with a multi-step movement plan $\dots$} \\
\textcolor{blue2}{I will make sure the joint plan is feasible for all robots under this configuration $\dots$} \\
\textcolor{blue2}{Let me finalize this plan: it coordinates robots in parallel and avoids unnecessary motion $\dots$} \\[0.25em]
\textcolor{blue2}{\textless think \textgreater} \\[0.25em]

{\ttfamily
[ \{ "Robot 0": "Move [0.75, 1.25, 0.09] True", $\dots$ \} ]
} \\

\midrule
\bottomrule
\end{tabularx}
\vspace*{-0.1in}
\end{table}

\begin{table}[h]
\small
\centering
\caption{Example reasoning trace for \icreplan planner.}
\label{tab:example-trace-icreplan}
\begin{tabularx}{0.65\textwidth}{Y}
\toprule \midrule
\textbf{Input}: \\
\textcolor{gray}{\textless observation\textgreater} \\
\textcolor{gray}{Object positions:} \\
\textcolor{gray}{\hspace*{2em}Object 0: [0.75, 2.75, 0.08] $\dots$} \\
\textcolor{gray}{\hspace*{2em}Object 1: [2.25, 2.75, 0.08] $\dots$} \\
\textcolor{gray}{Robot states:} \\
\textcolor{gray}{\hspace*{2em}Robot 0: base=[0.75, 2.25, 0.00], arm=[0.85, 2.35, 0.10] $\dots$} \\
\textcolor{gray}{\hspace*{2em}Robot 1: base=[2.25, 2.25, 0.00], arm=[2.15, 2.35, 0.10] $\dots$} \\
\textcolor{gray}{\textless /observation\textgreater}\\[0.25em]

\textbf{\icreplan Planner Reasoning}: \\
\textcolor{blue2}{\textless think\textgreater Alright, I will continue executing the current plan and only replan if the next step becomes invalid $\dots$} \\[0.25em]

{\ttfamily
[
\{ "Robot 0": "Move [1.50, 2.75, 0.10] True",
   "Robot 1": "Move [2.25, 2.25, 0.09] False" \} $\dots$
] 
}\\[0.25em]

\textcolor{gray}{\textless observation\textgreater} \\
\textcolor{gray}{Object positions: $\dots$} \\
\textcolor{gray}{Robot states: $\dots$} \\
\textcolor{gray}{Execution feedback:} \\
\textcolor{gray}{\hspace*{2em}Previous plan: \{ "Robot 0": "Move [1.50, 2.75, 0.10] True" $\dots$} \\
\textcolor{gray}{\hspace*{2em}FAIL: collision predicted at [1.50, 2.75, 0.10] $\dots$} \\
\textcolor{gray}{\textless /observation\textgreater}\\[0.25em]

\textcolor{blue2}{\textless think\textgreater Given this feedback, I need to re-analyze the environment and generate a new next step $\dots$} \\
\textcolor{blue2}{Let me finalize this updated plan by first clearing the block $\dots$}\\[0.25em]

{\ttfamily
\{ "Robot 1": "Move [2.50, 2.75, 0.09] True",
   "Robot 0": "Move [0.75, 3.00, 0.10] True" \} $\dots$
} \\

\midrule
\bottomrule
\end{tabularx}
\caption{Example reasoning trace for \fullplan planner.}
\label{tab:example-trace-fullplan}

\vspace*{-0.1in}
\end{table}

\begin{table}[h]
\small
\centering
\vspace*{-0.1in}

\begin{tabularx}{0.65\textwidth}{Y}
\toprule \midrule
\textbf{Input}: \\

\textcolor{gray}{\textless observation\textgreater} \\
\textcolor{gray}{Object positions:} \\
\textcolor{gray}{\hspace*{2em}Object 0: [0.75, 2.75, 0.08] $\dots$} \\
\textcolor{gray}{\hspace*{2em}Object 1: [2.25, 2.75, 0.08] $\dots$} \\
\textcolor{gray}{Robot states:} \\
\textcolor{gray}{\hspace*{2em}Robot 0: base=[0.75, 2.25, 0.00], arm=[0.85, 2.35, 0.10] $\dots$} \\
\textcolor{gray}{\hspace*{2em}Robot 1: base=[2.25, 2.25, 0.00], arm=[2.15, 2.35, 0.10] $\dots$} \\
\textcolor{gray}{\textless /observation\textgreater}\\[0.25em]

\textbf{\ncreplan Planner Reasoning}: \\
\textcolor{blue2}{\textless think\textgreater Alright, let's break down the situation and plan the next step $\dots$} \\
\textcolor{blue2}{Let me finalize this step to make progress efficiently $\dots$}\\[0.25em]

{\ttfamily
\{ "Robot 0": "Move [0.75, 2.50, 0.09] False", $\dots$
} \\[0.25em]

\textbf{RePlan Input}: \\
\textcolor{gray}{\textless observation\textgreater} \\
\textcolor{gray}{Object positions: $\dots$} \\
\textcolor{gray}{Robot states: $\dots$} \\
\textcolor{gray}{Execution feedback:} \\
\textcolor{gray}{\hspace*{2em}Previous plan: \{ "Robot 0": "Move [0.75, 2.50, 0.09] False" $\dots$} \\
\textcolor{gray}{\hspace*{2em}FAIL: Robot 0 motion infeasible $\dots$} \\
\textcolor{gray}{\textless /observation\textgreater}\\[0.25em]

\textbf{\ncreplan Planner Reasoning}: \\
\textcolor{blue2}{\textless think\textgreater Alright, let's analyze this again and generate a new next-step plan $\dots$} \\
\textcolor{blue2}{Let me finalize this updated step by rerouting Robot 1 $\dots$}\\[0.25em]

{\ttfamily
\{ "Robot 0": "Move [0.75, 2.75, 0.10] True", $\dots$
} \\

\midrule
\bottomrule
\end{tabularx}
\caption{Example reasoning trace for \ncreplan planner.}
\label{tab:example-trace-ncreplan}

\vspace*{-0.1in}
\end{table}

\subsection{Detailed Prompts}\label{sec:app-prompt}
In this section, we list all the prompts involved in all experiments in this work. Specifically, we mainly list the prompt for Task Planner such as \fullplan prompt and the prompt for motion planner, as well as the promtp for reasoning behavior check experiment.

\paragraph{Prompt for Task Planner}

We list the prompt for \fullplan task planner in the following listing. For \replan-mode, the main difference is that it involves a separate paragraph description about utilizing the error feedback, which is listed below the \fullplan planner prompt.

\begin{minted}{text}
You are a central planner responsible for coordinating multiple robotic arms operating in a 3D grid-like environment. Your goal is to plan and execute efficient, collision-free movements to transport objects to their designated target positions while avoiding obstacles.

## Task Description:
*Task Representation:*
* Objective: Move all objects to their specified target locations safely and efficiently.
* Input: A detailed map state containing positions of robots, objects, targets, and obstacles.
* Output: A precise movement plan specifying each robot arm's actions for moving objects.

*Position Representation:*
* All positions (robots, objects, targets, obstacles) are given by their center coordinates, e.g., [0.55, 1.65, 0.08].
* Robots have a fixed base location and an extendable arm with a limited reach range.

*Reachability Rules:*
* Each robot arm can only move within an axis-aligned reach box centered at its base position:
  - Let dx = X - Base_X, dy = Y - Base_Y, dz = Z - Base_Z.
  - The arm may reach (X, Y, Z) only if:
    - -0.9 <= dx <= 0.9
    - -0.9 <= dy <= 0.9
    - 0.001 <= dz <= 0.8
* If a robot needs to move an object within its range and the arm is not aligned with the object, the robot should first move its arm to the object position. By aligning, the distance between object center and arm position is less than 0.1.
* When you plan a move, please follow these rules:
  - First check that the proposed target lies within the reach box.
  - If it does not, adjust your plan or reject that movement.
  - If the arm is not yet aligned with an object it needs to move and that object lies within the reach box, plan a preliminary move to position the arm aligned with the object before carrying it.

*Obstacles:*
* The environment may contain fixed obstacles that robot arms must avoid during movement. Obstacles are static and cannot be moved or altered.
* Each obstacle is defined by:
  - Type — one of: "box" or "pole".
  - Center position — [x, y, z] coordinates of the obstacle's center.
  - Size parameters:
    - Box: size = [half_x, half_y, half_z] (half extents along each axis).
    - Pole: size = [radius, half_height].
  - Height may also be provided as a full height; if both size and height are present, treat the obstacle as a solid volume covering that height.
* Collision Constraint:
  - The robot arm's path, including all intermediate positions and any carried object's swept volume, must not intersect or pass through any part of the obstacle's volume.
*Motion Planning Feasibility:*
* A separate motion planner exists to generate the detailed movement trajectories.
* Your task-level plan should only propose reasonable target positions that a motion planner could feasibly connect, especially around obstacles; avoid targets that would require the arm or carried object to pass through obstacles, and sequence actions if simultaneous moves would be infeasible.

## How to Generate Your Response:
Your response must **clearly indicate your thinking process** enclosed in <think> and </think> tags, followed by the generated step of the movement plan.

*Thinking:*
* Clearly describe your analysis and decisions from a first-person perspective.
* Identify potential collisions explicitly (including obstacle collisions) and explain how you avoid them.
* Highlight your reasoning for movement choices, considering efficiency and collision avoidance.
* Include explicit checks and self-questioning in your thinking process.

*Movement Plan (Output):*
* Your generated step of the movement plan should be in markdown format and contain a JSON array, with robot names as keys and their movement instructions as values, structured as follows:
```json
[
  {
    "Robot 0": "Move [x, y, z] True",
    "Robot 1": "Move [x, y, z] False"
  },
  {
    "Robot 2": "Move [x, y, z] True"
  }
]
```
* *[x, y, z]* represent the target coordinates of the robot arm end effector.
* *True/False* indicates whether the robot moves an object (True) or simply moves its arm without carrying an object (False).
* One robot can only be moved once in each step, which means no repeated keys are allowed in the same step.
* Robots without actions in the current step should not be included.
Ensure your final step completes the objective of placing all objects at their target positions, and your plan forms a valid JSON array.

## Collision Avoidance Rules:
Your plan must strictly avoid collisions, as follows:
* Robot-Robot Collision
  * Two robot arms cannot occupy the same position at the end of a move.
  * Their paths must not cross or share any point during the move; if unsure, sequence the motions instead of parallelizing.
* Object-Object Collision
  * Two objects cannot occupy the same (x, y, z) at any time.
  * If you move more than one object at once, they must have different drop-off points and non-crossing straight-line paths.
* Robot-Object Collision
  * An arm's path must not sweep through any object it isn't carrying.
  * Before moving, confirm the trajectory does not pass over another object's position.
* Robot-Obstacle Collision
  * An arm's path and any carried object's swept volume must not intersect obstacles.
  * If an obstacle blocks a direct move, plan an intermediate waypoint or a different robot/sequence.

## Plan Efficiency Considerations:
* Each step of your plan involves simultaneous robot arm movements from their current positions to specified target positions.
* Each robot arm moves at a constant speed of 0.5 units/time.
* The duration of each step is determined by the longest single-arm movement within that step.
* The total execution time is the sum of all individual step durations.
* You should aim to minimize total execution time while ensuring collision-free movements and successful object placements.

Given the information above, now consider the following environment:
Input:
{mapstate}

Generate the full plan for moving these robots.
\end{minted}
\begin{minted}{text}
*Replanning With Failure Feedback:*
* You may be given failure feedback from previous failed attempts, which contains the last-step action plan that leads to any error like collision or infeasible move and the specifica reason for that failure.
* Use that feedback to avoid repeating the same error step and error type, and adjust the plan accordingly before producing the new plan.

*Failure Feedback Format (Example):*
```text
Failed attempt (last-step plan):
{
  "Robot 0": "Move [x, y, z] True",
  "Robot 1": "Move [x, y, z] False"
}
Failure reasons:
- Collision between Robot 0 and Robot 1
- Collision between Robot 0 and obstacle at (x, x)
```
\end{minted}

\paragraph{Prompt for Motion Planner}

We list the prompt for motion planner in the following listing.
\begin{minted}{text}
You are a motion planner responsible for controlling a single robotic arm to move an object (or the arm alone) from a start point to a target position safely. Your task is to generate a collision-free, waypoint-based movement plan that respects reachability and obstacle constraints.

## Task Description:
*Task Representation:*
* Objective: Move the robot arm (and any carried object) from the given start point to the specified target position without collisions.
* Input: Start point, target position, and observation of the adjacent environment (objects and obstacles near the path).
* Output: A precise movement plan specifying the robot arm's waypoint trajectory.

*Position Representation:*
* All positions (robot base, start, target, objects, obstacles) are given by their center coordinates, e.g., [0.55, 1.65, 0.08].
* The robot has a fixed base location and an extendable arm with a limited reach range.

*Reachability Rules:*
* Each robot arm can only move within a sphere centered at its base position:
  - Let d = sqrt((X - Base_X)**2 + (Y - Base_Y)**2 + (Z - Base_Z)**2).
  - The arm may reach (X, Y, Z) only if d < 0.8.
* If the robot needs to move an object within its range and the arm is not aligned with the object, the robot should first move its arm to the object position. By aligning, the distance between object center and arm position is less than 0.1.
* When you plan a move, please follow these rules:
  - First check that the proposed waypoint lies within the reach sphere.
  - If it does not, adjust your plan or reject that movement.
  - If the arm is not yet aligned with an object it needs to move and that object lies within reach, plan a preliminary waypoint to align before carrying it.

*Obstacles:*
The environment may contain fixed obstacles that robot arms must avoid during movement. Obstacles are static and cannot be moved or altered.
Each obstacle is defined by:
* Type - one of:
  * "box" - a solid rectangular prism
  * "pole" - a solid vertical cylinder
* Center position - [x, y, z] coordinates of the obstacle's center in the environment.
* Size parameters:
  * Box:
    * length - size along the X-axis
    * width - size along the Y-axis
    * height - vertical height of the box
  * Pole:
    * radius - radius of the circular base
    * height - vertical height of the pole

*Collision Constraint:*
The robot arm's path, including all intermediate waypoints and any carried object's swept volume, must not intersect or pass through any part of the obstacle's volume.
For poles, this means no path point may enter the radius (plus a safety margin if needed).
For boxes, this means no path segment may pass through the rectangular footprint (length x width) at any height.

*Waypoint Trajectory Rules:*
Every motion should be represented as a sequence of waypoints, not just a start and end position. Below are the detailed rules for generating these waypoints:
* The arm should only move to a new waypoint if that waypoint satisfies the reachability constraint (d < 0.8 from the base).
* Each movement step in your plan must include the full waypoint list for that movement.
* You may assume the robot arm moves smoothly between waypoints, with constant speed, and each step can include multiple waypoints.

## How to Generate Your Response:
Your response should contain the generated movement plan only.

*Movement Plan (Output):*
* Your generated movement plan should be in markdown format and contain a JSON object, with each entry as a dictionary indicating one step. Use a single robot key and include waypoints only:
```json
{
    "waypoints": [[x1, y1, z1], [x2, y2, z2]]
}
```
Ensure your final step reaches the target position, and your plan forms a valid JSON.

## Collision Avoidance Rules:
Your plan must strictly avoid collisions, as follows:
* Robot-Object Collision
  * The arm's path must not sweep through any object it isn't carrying.
  * Before moving, confirm the trajectory does not pass over another object's position.
* Robot-Obstacle Collision
  * The arm's path and any carried object's swept volume must not intersect obstacles.
  * If an obstacle blocks a direct move, plan an intermediate waypoint.
Given the information above, now consider the following environment:
Input:
{mapstate}
{motion movement request}
Generate the motion plan for the single robot
\end{minted}

\paragraph{Prompt for Task Plan Reasoning Trace Synthesis}

We list the prompt for Task planner reasoning trace generation in the following listing.
\begin{minted}{text}
You are required to **assume the role of a central planner**. Your task is to simulate the step-by-step thinking process that logically leads you to the provided ground-truth movement plan.

Your thinking should be presented from a **first-person perspective**, clearly demonstrating your internal reasoning process of planning, validating and adjusting to avoid collision, and planning decisions.

## Requirement for your generated first-person thinking:
1. **First-Person Perspective**: Write your internal thoughts as if you are personally making the decisions:
   - Use phrases like "Let me see...", "Wait, is that correct?", "I should check collisions first...", "Can I parallel two robot movements to make the plan more efficient?"
   - Demonstrate real-time analysis and potential hesitations or reconsiderations.
2. **Thinking Process with `<think>` Tags**:
   - Enclose your entire reasoning sequence in `<think>` ... `</think>` tags.
   - Make sure you have explicit checks, e.g. collision checks, reachability, and confirmations of correctness. You can start the explicit checks with "Wait", "Hmm", "let me check", etc.
   - Make sure to pose questions to yourself, and then answer them. Show how you arrive at each movement decision.
   - You must include multiple explicit checks and self-questioning in your thinking process.
3. You should not reveal any information about the guidance here that you are synthesizing a thinking process.

Below is the detailed task description. You can learn the rules for the task from these descriptions.

## Task Description:
You are a central planner responsible for coordinating multiple robotic arms operating in a 3D grid-like environment. Your goal is to plan and execute efficient, collision-free movements to transport objects to their designated target positions while avoiding obstacles.

*Task Representation:*
* Objective: Move all objects to their specified target locations safely and efficiently.
* Input: A detailed map state containing positions of robots, objects, targets, and obstacles.
* Output: A precise movement plan specifying each robot arm's actions for moving objects.

*Position Representation:*
* All positions (robots, objects, targets, obstacles) are given by their center coordinates, e.g., [0.55, 1.65, 0.08].
* Robots have a fixed base location and an extendable arm with a limited reach range.

*Reachability Rules:*
* Each robot arm can only move within an axis-aligned reach box centered at its base position:
  - Let dx = X - Base_X, dy = Y - Base_Y, dz = Z - Base_Z.
  - The arm may reach (X, Y, Z) only if:
    - -0.9 <= dx <= 0.9
    - -0.9 <= dy <= 0.9
    - 0.001 <= dz <= 0.8
* If a robot needs to move an object within its range and the arm is not aligned with the object, the robot should first move its arm to the object position. By aligning, the distance between object center and arm position is less than 0.1.
* When you plan a move, please follow these rules:
  - First check that the proposed target lies within the reach box.
  - If it does not, adjust your plan or reject that movement.
  - If the arm is not yet aligned with an object it needs to move and that object lies within the reach box, plan a preliminary move to position the arm aligned with the object before carrying it.

*Obstacles:*
* The environment may contain fixed obstacles that robot arms must avoid during movement. Obstacles are static and cannot be moved or altered.
* Each obstacle is defined by:
  - Type — one of: "box" or "pole".
  - Center position — [x, y, z] coordinates of the obstacle’s center.
  - Size parameters:
    - Box: size = [half_x, half_y, half_z] (half extents along each axis).
    - Pole: size = [radius, half_height].
  - Height may also be provided as a full height; if both size and height are present, treat the obstacle as a solid volume covering that height.
* Collision Constraint:
  - The robot arm’s path, including all intermediate positions and any carried object’s swept volume, must not intersect or pass through any part of the obstacle’s volume.

## How to Generate Your Response:
Your response must **clearly indicate your thinking process** enclosed in <think> and </think> tags, followed by the generated step of the movement plan.

*Thinking:*
* Clearly describe your analysis and decisions from a first-person perspective.
* Identify potential collisions explicitly (including obstacle collisions) and explain how you avoid them.
* Highlight your reasoning for movement choices, considering efficiency and collision avoidance.

*Movement Plan (Output):*
* Your generated step of the movement plan should be in markdown format and contain a JSON array, with robot names as keys and their movement instructions as values, structured as follows:
```json
[
  {
    "Robot 0": "Move [x, y, z] True",
    "Robot 1": "Move [x, y, z] False"
  },
  {
    "Robot 2": "Move [x, y, z] True"
  }
]
```
* *[x, y, z]* represent the target coordinates of the robot arm end effector.
* *True/False* indicates whether the robot moves an object (True) or simply moves its arm without carrying an object (False).
* One robot can only be moved once in each step, which means no repeated keys are allowed in the same step.
* Robots without actions in the current step should not be included.
Ensure your final step completes the objective of placing all objects at their target positions, and your plan forms a valid JSON array.

## Collision Avoidance Rules:
Your plan must strictly avoid collisions, as follows:
* Robot-Robot Collision
  * Two robot arms cannot occupy the same position at the end of a move.
  * Their paths must not cross or share any point during the move; if unsure, sequence the motions instead of parallelizing.
* Object-Object Collision
  * Two objects cannot occupy the same (x, y, z) at any time.
  * If you move more than one object at once, they must have different drop-off points and non-crossing straight-line paths.
* Robot-Object Collision
  * An arm's path must not sweep through any object it isn't carrying.
  * Before moving, confirm the trajectory does not pass over another object's position.
* Robot-Obstacle Collision
  * An arm's path and any carried object’s swept volume must not intersect obstacles.
  * If an obstacle blocks a direct move, plan an intermediate waypoint or a different robot/sequence.

## Plan Efficiency Considerations:
* Each step of your plan involves simultaneous robot arm movements from their current positions to specified target positions.
* Each robot arm moves at a constant speed of 0.5 units/time.
* The duration of each step is determined by the longest single-arm movement within that step.
* The total execution time is the sum of all individual step durations.
* You should aim to minimize total execution time while ensuring collision-free movements and successful object placements.

## Example Environment and Ground-Truth Plan:
Below is an example scenario and its ground-truth solution:

```text
{Environment and GT plan}
```

With the above information clearly provided, please start by explicitly presenting your first-person reasoning for the whole plan enclosed in <think></think> tags. Make sure you include explicit checks and self-questioning in your thinking process. Your reasoning should be clear and easy to follow, as if you are explaining it to someone else. Limit your thinking length within 2000 tokens.
\end{minted}

\paragraph{Prompt for Reasoning Behavior Check}

We list the prompt for reasoning behavior check count in the following listing.
\begin{minted}{text}
# Prompt for reachability check count
You are given a planner response. Count the reachability checks.
Point to each appearance by quoting a short phrase from the response.
Return a JSON object with the total count and the list of appearances.

## Example Texts (Reachability Checks)
- "Is target [1.65, 0.55, 0.08] within Robot 0's reach box (dx, dy in [-0.9, 0.9], dz in [0.001, 0.8])?"
- "dx=1.05 exceeds 0.9, so Robot 2 cannot reach this target."
- "The object is within the reach box; I can align the arm before carrying it."

## Output Format
{
  "behavior": "reachability_check",
  "count": 2,
  "appearances": ["short quote 1", "short quote 2"]
}


# Prompt for robot-robot collision check count
You are given a planner response. Review the response and count robot-robot collision checks.
Point to each appearance with a brief quoted phrase.
Return a JSON object with the total count and the list of appearances.

## Example Texts (Robot-Robot Collision Checks)
- "If Robot 0 and Robot 1 move simultaneously, their paths would intersect; I will sequence them."
- "Both arms would end at the same position, which is a robot-robot collision."
- "Their paths may cross, so I avoid parallel moves and move one arm at a time."

## Output Format
{
  "behavior": "robot_robot_collision_check",
  "count": 2,
  "appearances": ["short quote 1", "short quote 2"]
}


# Prompt for robot-obstacle collision check count
You are given a planner response. Count robot-obstacle collision checks in the response.
For each check, quote a short phrase that shows where it appears.
Return a JSON object with the total count and the list of appearances.

## Example Texts (Robot-Obstacle Collision Checks)
- "The arm's swept volume would intersect the box obstacle at center [1.10, 0.75, 0.10], so I avoid a direct move."
- "A straight-line move passes through the pole volume; I choose a waypoint around it."
- "The obstacle blocks the path, so I plan an intermediate target to keep clearance."

## Output Format
{
  "behavior": "robot_obstacle_collision_check",
  "count": 2,
  "appearances": ["short quote 1", "short quote 2"]
}

# Prompt for error recovery check count
You are given a planner response. Count the corrections of previous errors.
For each correction, show where it appears with a short quote.
Return a JSON object with the total count and the list of appearances.

## Example Texts (Corrections of Previous Errors)
- "I previously ignored the obstacle volume; correcting that by rerouting."
- "My last plan caused a robot-robot collision, so I change the sequence."
- "Earlier I chose the wrong robot; I fix that based on reachability."

## Output Format
{
  "behavior": "correction_previous_errors",
  "count": 2,
  "appearances": ["short quote 1", "short quote 2"]
}
\end{minted}

\end{document}